\pdfoutput=1

\documentclass[11pt]{article}

\usepackage{amsmath,amsfonts,bm}

\newcommand{\grad}[2]{\nabla\ell(#1;#2)}
\newcommand{\update}[2]{\Gamma(#1, #2)}
\newcommand{\infadam}{{\textrm{Inf}_{\text{Adam}}}}

\def\eqref#1{equation~\ref{#1}}

\def\1{\bm{1}}

\def\vtheta{{\bm{\theta}}}

\def\vm{{\bm{m}}}

\def\vt{{\bm{t}}}

\def\vv{{\bm{v}}}

\def\mA{{\bm{A}}}

\DeclareMathAlphabet{\mathsfit}{\encodingdefault}{\sfdefault}{m}{sl}
\SetMathAlphabet{\mathsfit}{bold}{\encodingdefault}{\sfdefault}{bx}{n}

\def\gD{{\mathcal{D}}}

\def\gM{{\mathcal{M}}}

\def\gT{{\mathcal{T}}}

\def\gV{{\mathcal{V}}}

\DeclareMathOperator*{\argmax}{arg\,max}

\usepackage[utf8]{inputenc} %
\usepackage[T1]{fontenc}    %
\usepackage{hyperref}       %
\usepackage{url}            %
\usepackage{booktabs}       %
\usepackage{amsfonts}       %
\usepackage{nicefrac}       %
\usepackage{microtype}      %
\usepackage{xcolor}         %
\usepackage{standalone}
\usepackage{latexsym}
\usepackage{amsmath}
\usepackage{amssymb}
\usepackage{amsthm}
\usepackage{natbib}   
\usepackage{graphicx}
\usepackage{subcaption}
\usepackage{array}
\usepackage{tabu}
\usepackage{makecell}
\usepackage{paralist}
\usepackage{cases}
\usepackage{diagbox}
\usepackage{enumitem}
\usepackage{soul}
\usepackage{multirow}
\usepackage{verbatim}
\usepackage{tabulary}
\usepackage{booktabs}
\usepackage{tabularx}
\usepackage[mathscr]{euscript}
\usepackage{mathtools}
\usepackage{algorithm}
\usepackage{algpseudocode}
\usepackage{stmaryrd}
\usepackage{tikz-dependency}
\usetikzlibrary{automata,decorations.markings,arrows,positioning,matrix,calc,patterns,angles,quotes,calc}
\usepackage{adjustbox}
\usepackage{tabularx}
\usepackage{xspace}
\usepackage{tabulary}
\usepackage{afterpage}
\usepackage{bm}
\usepackage{color}
\usepackage{graphicx}
\usepackage{slashbox}
\usepackage[toc,page]{appendix}
\usepackage{makecell}
\usepackage{boldline}
\usepackage[shortcuts]{extdash}  %

\usepackage{blindtext}
\usepackage{graphicx}
\usepackage{capt-of}
\usepackage{booktabs}
\usepackage{varwidth}
\usepackage{pifont}%
\usepackage{wrapfig}

\usepackage{listings}

\usepackage{pythonhighlight}

\usepackage[normalem]{ulem}
\useunder{\uline}{\ul}{}

\definecolor{orange}{rgb}{1,0.5,0}
\definecolor{mdgreen}{rgb}{0.05,0.6,0.05}
\definecolor{mdblue}{rgb}{0,0,0.7}
\definecolor{dkblue}{rgb}{0,0,0.5}
\definecolor{dkgray}{rgb}{0.3,0.3,0.3}
\definecolor{slate}{rgb}{0.25,0.25,0.4}
\definecolor{gray}{rgb}{0.5,0.5,0.5}
\definecolor{ltgray}{rgb}{0.7,0.7,0.7}
\definecolor{purple}{rgb}{0.7,0,1.0}
\definecolor{lavender}{rgb}{0.65,0.55,1.0}

\definecolor{mypurple}{RGB}{111,61,121}
\definecolor{myblue}{RGB}{46,88,180}
\definecolor{myred}{RGB}{181,68,106}
\definecolor{myyellow}{RGB}{204,143,55}

\newcommand{\term}[1]{\textbf{#1}} %

\DeclareSymbolFont{extraup}{U}{zavm}{m}{n}
\DeclareMathSymbol{\vardiamond}{\mathalpha}{extraup}{87}

\newcolumntype{L}[1]{>{\raggedright\let\newline\\\arraybackslash\hspace{0pt}}m{#1}}
\newcolumntype{C}[1]{>{\centering\let\newline\\\arraybackslash\hspace{0pt}}m{#1}}
\newcolumntype{R}[1]{>{\raggedleft\let\newline\\\arraybackslash\hspace{0pt}}m{#1}}

\theoremstyle{definition}

\theoremstyle{remark}

\algrenewcommand{\algorithmiccomment}[1]{\leavevmode$\triangleright$ #1}

\setul{1pt}{.4pt}

\newcommand*{\name}{\textsc{BIDS}\xspace}
\newcommand*{\cott}{\textsc{COT}\xspace}
\newcommand*{\dolly}{\textsc{DOLLY}\xspace}
\newcommand*{\flan}{\textsc{FLAN V2}\xspace}
\newcommand*{\oasst}{\textsc{OPEN ASSISTANT 1}\xspace}

\DeclareFixedFont{\ttb}{T1}{txtt}{bx}{n}{12} %
\DeclareFixedFont{\ttm}{T1}{txtt}{m}{n}{12}  %

\newcommand{\StateVSpace}{\vspace{0.25em}}

\setlist[enumerate]{leftmargin=*, itemsep=1mm, parsep=1mm, topsep=1mm, partopsep=1mm}
\setlist[itemize]{leftmargin=*, itemsep=1mm, parsep=1mm, topsep=1mm, partopsep=1mm}

\usepackage[final]{acl}

\usepackage{times}
\usepackage{latexsym}

\usepackage[T1]{fontenc}

\usepackage[utf8]{inputenc}

\usepackage{microtype}

\usepackage{inconsolata}

\usepackage{graphicx}

\title{Improving Influence-based Instruction Tuning Data Selection for Balanced Learning of Diverse Capabilities}

\author{Qirun Dai$^{1,2}$\thanks{Work done during internship at UIUC.}, Dylan Zhang$^{1}$, Jiaqi W. Ma$^{1}$, Hao Peng$^{1}$\\
$^1$University of Illinois Urbana-Champaign
$^2$The University of Chicago\\
\texttt{qirundai@uchicago.edu}\quad
\texttt{\{shizhuo2,jiaqima,haopeng\}@illinois.edu}\\
}

\usepackage{xcolor}

\begin{document}
\maketitle
\begin{abstract}
Selecting appropriate training data is crucial for instruction fine-tuning of large language models (LLMs), which aims to (1) elicit strong capabilities, and (2) achieve balanced performance across different tasks. 
Influence-based methods show promise in achieving (1), by estimating the contribution of each training example to the model's predictions, but often struggle with (2). 
Our systematic investigation reveals that this underperformance can be attributed to an inherent bias,
where some tasks intrinsically have greater influence than others. 
As a result, data selection is often biased towards these tasks, not only hurting the model's performance on others but also, counterintuitively, harming performance on these high-influence tasks themselves.
To address this, we propose \name, a \emph{\textbf{B}alanced and \textbf{I}nfluential \textbf{D}ata \textbf{S}election} algorithm. 
\name first normalizes influence scores of the training data,
and then iteratively chooses the training example with the highest influence on the most underrepresented task. 
Experiments with both Llama-3 and Mistral-v0.3 on seven benchmarks spanning five diverse capabilities show that 
\name consistently outperforms \emph{both} state-of-the-art influence-based  algorithms and other non-influence-based frameworks. Surprisingly, training on a 15\% subset selected by \name can even outperform full-dataset training with a much more balanced performance. 
Our analysis highlights the importance of both instance-level normalization and iterative optimization of selected data for balanced learning of diverse capabilities.
\end{abstract}

\section{Introduction}\label{sec:intro}

Instruction tuning plays a crucial role in eliciting strong capabilities from LLMs. 
Typically, a pretrained LLM is finetuned on a mixture of different instruction datasets, to learn to follow human instructions and achieve strong and balanced performance on a wide range of capabilities~\citep{ouyang2022instructgpt, touvron2023llama, dubey2024llama, jiang2023mistral}.
The importance of data quality for instruction tuning~\citep{zhou2024lima} has spawned many works on data selection~\citep{cao2023instruction, chen2023alpagasus, liu2024what}.
Among them, influence-based methods, which estimate each training example's influence on the model's prediction on a downstream task~\citep{koh2017understanding, pruthi2020estimating}, have shown promise.
Thanks to recent advances, they have also been scaled to LLM-level computations and demonstrated strong potential in selecting high-quality instruction tuning data~\citep{xia2024less, choe2024your, yu2024mates}. 

However, influence estimation methods are typically designed to measure the data influence for a single task~\citep{koh2017understanding, pruthi2020estimating}. 
In this study, we demonstrate that existing influence-based data selection algorithms~\citep{xia2024less} struggle to balance capabilities across diverse tasks, which is crucial in real-world applications.\footnote{E.g., it is desirable for a coding agent to faithfully follow user instructions and perform complex reasoning.} Specifically, our analysis reveals that the influence scores for some tasks can be larger than others' by several magnitudes.
This can introduce systematic biases in the data selection process when cross-task influence scores are directly compared, as done in many existing works~\citep{yin2025computeconstrained, albalak2024a}. 
This leads to a couple of pitfalls.
First, biasing towards some tasks hurts the model's performance on others, making it more challenging for LLMs to achieve balanced capabilities.
Second, perhaps counterintuitively, it may even hurt the model's performance on the very task that the data is biased towards.
These issues call for an influence-based selection algorithm designed for training LLMs to achieve balanced capabilities across diverse tasks.

\name, our proposed algorithm, addresses these challenges with two key designs.
Given a training dataset to select from and a validation dataset representing diverse target tasks,
we describe the influence pattern between them with a matrix,
where each column consists of the influence scores of different training examples on a specific validation instance.
\name first applies column-wise normalization to this matrix, thus setting the influence for different validation instances on the same scale.
Then, in contrast to prior methods that simply select instances with the highest influence values, \name applies an iterative selection algorithm. At each iteration, this algorithm compares the influence of each candidate with the average influence of those already selected, and selects the one that can provide the largest improvement for the current selected subset.
If the current dataset falls short in influence on certain validation instances, then our algorithm will intuitively favor training examples that have high influence on the specific tasks represented by these validation data,
and thus promotes balanced multi-task learning.

We evaluate \name on UltraInteract~\citep{yuan2024advancing}, a strong instruction tuning dataset suite, with Llama-3-8B~\citep{dubey2024llama} and Mistral-7B-v0.3~\citep{jiang2023mistral}.
Across seven tasks spanning five capabilities including coding, math, logical inference, world knowledge and instruction following, \name consistently outperforms both influence- and non-influence-based baselines.
Surprisingly, a 15\% subset selected by \name even outperforms full-dataset training in average performance. 
Further analysis highlights the importance of the key design choices of \name, and reveals completion length as a potential factor for biased influence.
The contributions of this work include: 
(1) Through systematic analysis, we identify inherent biases in current influence-based instruction tuning data selection algorithms.
(2) We propose \name to address this issue and promote balanced learning of multiple capabilities.
(3) Extensive experiments confirm the strong and consistent performance of \name, and provide valuable insights on what makes a balanced set of influential data.

\section{Background and Preliminaries}\label{sec:background}

\begin{table*}[t]
  \centering
  \resizebox{0.95\linewidth}{!}{
  \begin{tabular}{@{}cccccccccc@{}}
    \toprule
    \multirow{2}{*}{\textbf{Budget}} &
      \multirow{2}{*}{\textbf{Method}} &
      \multicolumn{2}{c}{\textbf{Coding}} &
      \textbf{Logic} &
      \textbf{Knowledge} &
      \multicolumn{2}{c}{\textbf{Math}} &
      \textbf{Ins-Following} &
      \multirow{2}{*}{\textbf{Macro Avg}} \\ \cmidrule(lr){3-9}
                          &        & HumanEval     & MBPP          & BBH           & MMLU          & GSM-Plus      & MATH          & IFEval        &               \\ \midrule
    \multirow{2}{*}{5\%}  & Random & 43.5          & 48.9          & \textbf{64.8} & 64.9          & 41.5          & 22.5          & 18.1          & 43.4          \\
                          & LESS   & \textbf{43.9} & \textbf{50.7} & 62.7          & \textbf{65.1} & \textbf{42.5} & \textbf{22.6} & \textbf{19.7} & \textbf{43.9} \\ \midrule
    \multirow{2}{*}{10\%} & Random & \textbf{47.8} & 50.6          & \textbf{65.0}   & \textbf{64.9} & 43.9          & 24.0            & 17.8          & \textbf{44.9} \\
                          & LESS   & 44.7          & \textbf{51.3} & 62.0            & 64.7          & \textbf{44.6} & \textbf{24.3} & \textbf{19.3} & 44.4          \\ \midrule
    \multirow{2}{*}{15\%} & Random & \textbf{48.7} & \textbf{51.9} & \textbf{65.2} & \textbf{65.1} & \textbf{45.6} & \textbf{25.0}   & 18.8          & \textbf{45.7} \\
                          & LESS   & 46.5          & 51.0            & 63.2          & 64.6          & 44.9          & 24.9          & \textbf{21.2} & 45.2          \\ \bottomrule
    \end{tabular}
  }
  \caption{
  Compared with the random baseline, 
  LESS shows weak macro-average and unbalanced task-specific performance, which suggests a potential bias in its selection process.
  }
  \label{tab:initial_results}
\end{table*}

\paragraph{Influence-based instruction tuning data selection.}
Estimating the influence of training examples on model predictions is critical for understanding model behavior.
Traditional methods, including retraining-based~\citep{ghorbani2019data, ilyas2022datamodels, park2023trak} and gradient-based~\citep{koh2017understanding, pruthi2020estimating} ones, %
have proven effective but are computationally prohibitive when scaling to LLMs, 
as they require either retraining on a large number of subsets, or computing at least a forward and backward pass for each training example in order to obtain its gradient~\citep{hammoudeh2024training, ko2024mirrored}.
Several recent advances seek to address these challenges by extending gradient-based approaches to scale more efficiently.
Given a large training dataset to select from and a validation set representing certain targeted capability, LESS~\citep{xia2024less} models the influence between each pair of training and validation examples through LoRA-based low-dimensional gradient similarity, and then selects training points with the highest influence on the validation set. LOGRA~\citep{choe2024your} leverages a low-rank gradient projection algorithm to further improve the efficiency of this process.
MATES~\citep{yu2024mates} formulates the pointwise data influence between each training point and the whole validation set, and uses a small model to learn this pointwise influence. %

Upon closer inspection, these LLM-scale influence-based data selection methods share a similar problem formulation. They all need a validation set to represent a targeted data distribution
and require pointwise
data influence between pair of training and validation instances.
We aim to extend this 
to more general-purpose instruction tuning, where the model simultaneously learns multiple diverse capabilities that may require training data from drastically different distributions.
Concretely, since only LESS directly targets instruction tuning among the three LLM-scale approaches, we ground our study on the specific formulation of LESS. But we emphasize that due to the highly similar influence modeling patterns
, the results of our work should also provide useful insights for other influence-based selection methods.

\paragraph{Problem Setup and Notations.}

Assume an instruction tuning dataset $\gD$, a validation dataset $\gV$, which spans $m$ diverse tasks that we want to optimize the LLM's performance for: $\gV=\gV_1\cup \dots\cup \gV_m$, and an \textbf{influence estimation method} that can compute the influence of each training example on each validation instance. We first compute the influence score between each pair of training and validation data,
yielding a $|\gD| \times |\gV|$ matrix $\mA$. 
Each row of $\mA$ corresponds to an individual training example, and each column a validation instance.
Element $\mA_{ij}$ indicates the influence of $i$-th example from $\gD$ on $j$-th instance from $\gV$. 
We dub $\mA$ an \term{Attribution Matrix (AM)} as it reveals the overall attribution pattern from the training set to all target tasks, and each row \hypertarget{notation:mA_i}{$\mA_i$} the \textbf{Influence Distribution} of the $i$-th training example.

Our goal is to design a \textbf{data selection algorithm} that can effectively 
select a subset $\gT$ from $\gD$ with size under a pre-defined budget, based on the information presented in $\mA$.
Finetuning the LLM on $\gT$ is supposed to achieve strong and balanced performance on all targeted tasks.
The evaluation tasks are specifically chosen to have minimal overlap in terms of the capabilities they benchmark.
The size of validation set for each task is also kept the same to avoid bias in the selection process.

\section{Existing Influence-based Selection Fails at Balancing Diverse Tasks}\label{sec:analysis}

We first show that LESS leads to suboptimal and unbalanced performance in 
general-purpose instruction tuning. 
This is quantitatively revealed by our analysis framework, which identifies inherent biases in the scale of influence values across different tasks.
Insights drawn in this section pave the way for the design choices of \name in \S\ref{sec:bids}.

\paragraph{Setting.} 
In this section, we use Llama-3-8B~\citep{dubey2024llama} as the base model for both influence estimation and evaluation of selected datasets.
For the instruction dataset to select from, we use UltraInteract~\citep{yuan2024advancing}, a state-of-the-art large-scale %
dataset designed to enhance diverse reasoning capabilities, including mathematical reasoning, coding, and general logical inference.
We also follow the evaluation setup of \citet{yuan2024advancing}, with seven datasets spanning five diverse capabilities. We use HumanEval~\citep{chen2021evaluating} and MBPP~\citep{austin2021program} for coding, GSM-Plus~\citep{li2024gsm} and MATH~\citep{hendrycks2021measuring} for math, and BigBench-Hard (BBH)~\citep{suzgun2022challenging} for general logical inference. We also use MMLU~\citep{hendrycks2020measuring} to assess the model's ability to understand world knowledge, and IFEval~\citep{zhou2023instruction} for fine-grained instruction-following ability. 
Appendix \ref{append:2_setup} contains more training and evaluation details.

We follow LESS for the \textbf{influence estimation method} throughout this work, with an equal number of validation instances sampled from each of the seven evaluation tasks to ensure fairness.
In this section, for the \textbf{data selection algorithm}, we also start with the \textbf{task-wise max} algorithm (Appendix \ref{append:3_baselines}) used by LESS.
For each training example, it first computes the mean influence over validation instances within the same task, followed by selecting training examples with the highest maximum influence across different tasks. 
We compare this algorithm against the random selection baseline.

\begin{figure*}[htbp]
  \centering
  \begin{minipage}{.48\textwidth}
    \input{figures/AID_all}
  \end{minipage}
  \hspace{0.02\textwidth} %
  \begin{minipage}{.48\textwidth}
    \input{figures/THI_task_p0.1}
  \end{minipage}
\end{figure*}

\paragraph{LESS fails to balance different capabilities (Table~\ref{tab:initial_results}).}
LESS shows substantial imbalance and variability in task-specific performance across different budgets. 
Although it consistently outperforms the random baseline in IFEval by more than $1.5\%$,
it underperforms in BBH by 2--3\%,
and shows no clear advantage in the remaining five tasks.
Moreover, with the increase of budget level, LESS is gradually outperformed by the random baseline in more tasks, leading to weaker macro-average under both 10\% and 15\% budgets.

The underperformance of LESS may stem from the fact that it is not designed for learning multiple diverse capabilities.
But given that the number of validation instances was kept equal for each task, it still suggests a potential inherent bias in the influence values,
which could skew the selection towards certain capabilities. If the overall influence on certain tasks is inherently higher, then the naive task-wise max algorithm will naturally prioritize them, possibly at the expense of others. 

In what follows, we aim to answer the following two questions: (1) whether influence values differ across tasks and to what extent, and (2) whether 
tasks with higher influence values receive more performance improvements.

\paragraph{What causes the imbalance of LESS?}

We first define two data analysis metrics:

\begin{itemize}[leftmargin=*, noitemsep, topsep=0pt, parsep=0pt, partopsep=0pt]
    \item \textbf{Average Influence Distribution (AID):} $\sum_{i=1}^{N}\mA_{i}/N$, is the average of influence distributions
    of all the training examples.
    This vector captures the overall influence pattern for all validation data, and is helpful to answering (1).
    \item \textbf{Task Frequency with Highest Influence (THI)} for a task $t$ is the number of selected examples that have the highest average influence on $t$. Distribution of THI targets (2) in identifying high-influence tasks for the selected subset.
\end{itemize}

\begin{figure*}[t]%
  \centering
  \includegraphics[width=\textwidth]{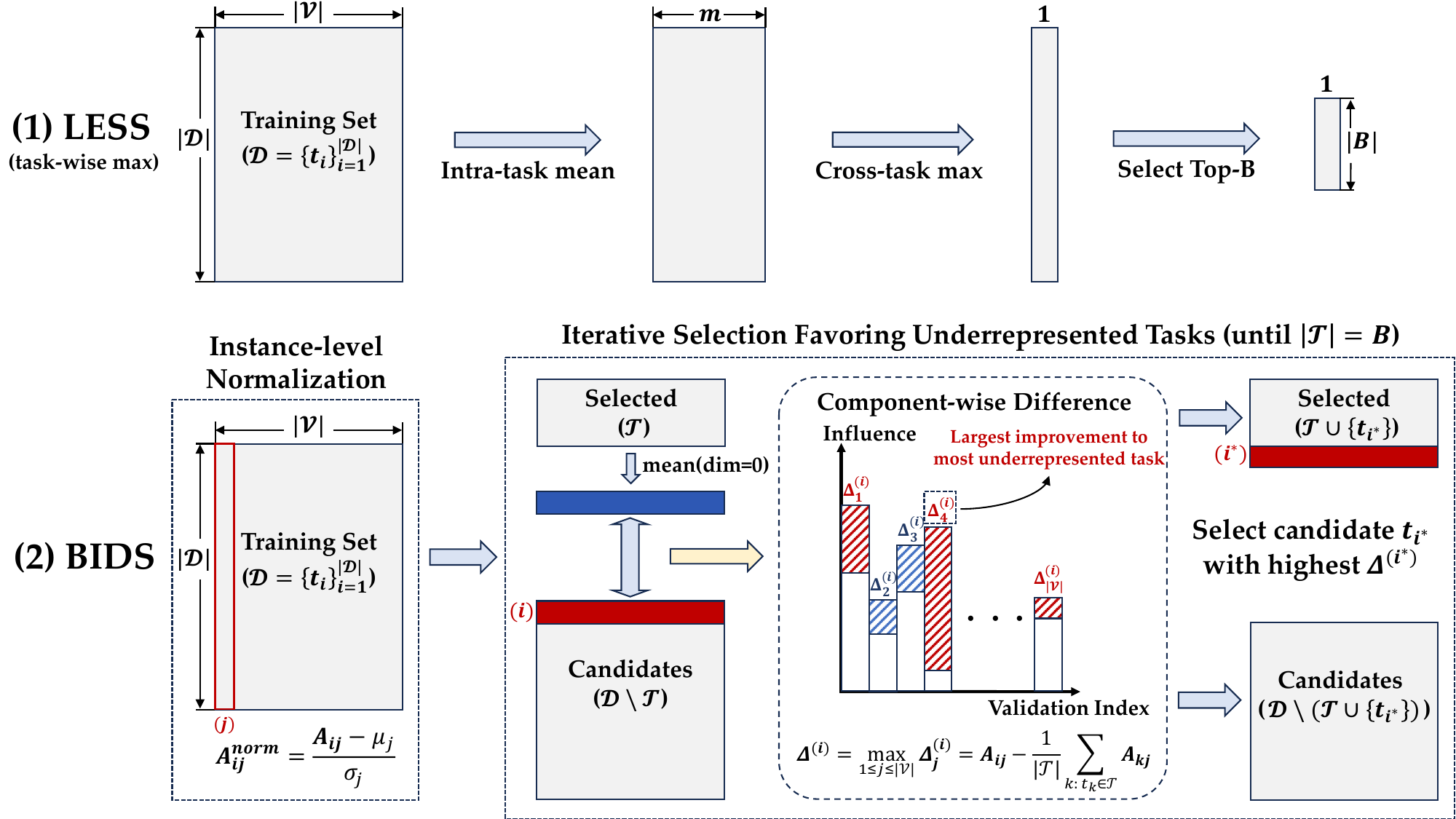}
  \caption{
  Comparison between \name and the task-wise max algorithm used by LESS. For convenience, we represent the training set $\gD$ with its Attribution Matrix (AM), in which the $i$-th row is the $|\gV|$-dimensional Influence Distribution of the $i$-th training example
  in $\gD$. 
  \name differs from LESS, which naively selects top-$B$ examples with the highest task-wise max utility scores, in mainly two aspects.
  \S\ref{sec:bids} contains a detailed walkthrough. 
  }
  \label{fig:pipeline}
\end{figure*}

Our UltraInteract AID analysis (Figure~\ref{fig:AID_all}) reveals discrepancies at both the task and instance levels.
Surprisingly, MMLU receives a substantially higher average influence than BBH, despite being more out-of-distribution for the training data.
Moreover, there are also significant disparities within the same task. 
For example, for IFEval, the gap between the highest and lowest instance-wise influence is greater than $0.0025$, 
2.5$\times$ larger than the globally highest instance-wise influence of $0.001$.
These results answer our question (1) by confirming that the scales of influence values differ significantly across various tasks.

Further, the THI analysis of LESS-selected data (Figure~\ref{fig:THI_task_p0.1}) confirms that these scale differences cause LESS to disproportionately favor certain tasks over others. 
Specifically, MMLU is most frequently identified as the most influential task, consistent with the observations in Figure~\ref{fig:AID_all} where MMLU has the highest task-level average influence. However, this does not translate into proportionally better performance---LESS frequently underperforms even the random baseline on MMLU. 
For other in-distribution tasks with high THI, such as MBPP, BBH, and GSM-Plus, LESS either consistently underperforms or shows no clear advantage.
These observations suggest that while high-influence tasks tend to have more examples in the selected training data, they do not necessarily offer greater room for performance improvement.
Besides, such biased sampling may hinder the learning of other capabilities.
Thus, we answer question (2) by concluding that the inherent difference in the scale of cross-task influence values is indeed a harmful bias, and can severely undermine the performance of LESS-selected data. 

\section{BIDS: Selecting Influential Data for Balanced Capability Learning}\label{sec:bids}

This section introduces \name to address the issues identified in \S\ref{sec:analysis}.
\name has two key design choices.

\paragraph{Instance-level normalization.}
This technique addresses the scale differences in influence values across different validation instances, by applying a column-wise normalization to the Attribution Matrix.
Specifically, for a validation instance $\vv_{j}$, the influence of each training example $\vt_{i}$ is normalized by
$\mA_{ij}^\text{norm} = (\mA_{ij} - \mu_j)/\sigma_j$,
where $\mu_{j}$ and $\sigma_{j}$ are the sample mean and standard deviation of all values in column $j$ of $\mA$. 
This normalization ensures the influence scores across different columns are comparable in scale. 
In other words, if two influence scores of different columns have similar intra-column rankings, then they should also have similar values.

\paragraph{Iterative selection favoring underrepresented tasks.}
We further propose an iterative greedy selection algorithm (Figure~\ref{fig:pipeline} and Algorithm~\ref{alg:bids})
to promote the balance across different capabilities.
It begins with an empty set.
In each iteration, it first computes the average influence distribution of the current selected subset $\gT$, denoted as $\mA_{\gT} \triangleq \frac{1}{|\gT|}\sum\limits_{k: \vt_k \in \gT}\mA_{k}$. 
Then it iterates through each training example $\vt_i$ in $\gD \setminus \gT$, and calculates a component-wise difference between $\mA_i$ and $\mA_{\gT}$. The utility $\Delta^{(i)}$ of candidate $\vt_i$ is then defined as the largest component of  $\mA_i - \mA_{\gT}$, and the candidate with the highest utility is selected.

\begin{algorithm}[htb]
    \small
    \caption{BIDS: Iterative Selection 
    }
    \label{alg:bids}
    \begin{algorithmic}[1]
    \State \textbf{Input:} $\gD$: 
    the set of all training examples;
    $\gV$: the set of validation examples;
    $B$: the number of examples to be selected; 
    $\mA \in \mathbb{R}^{|\gD| \times |\gV|}$: the Attribution Matrix between $\gD$ and $\gV$.
    
    \StateVSpace
    
    \State \textbf{Initialization:} $\gT = \varnothing$, $\gD = \{\vt_i\}_{i=1}^{|\gD|}$
    
    \StateVSpace
    
    \While{$|\gT| < B$}
        \StateVSpace
        \State $\Delta^{(i)} = \max\limits_{1 \le j \le |\gV|} \{ \mA_{ij} - \frac{1}{|\gT|}\sum\limits_{k \in \{k | \vt_k \in \gT\}} \mA_{kj} \}$
        \State $i^* = \argmax\limits_{i \in \{ i | \vt_i \in \gD \setminus \gT \}} \Delta^{(i)}$ 
        \StateVSpace
        \State $\gT = \gT \cup \{ \vt_{i^*} \}$
        \StateVSpace

    \EndWhile
    \StateVSpace
    \State \textbf{Return:} $\gT$: selected training examples.
    \end{algorithmic}
\end{algorithm}

In other words, \name prioritizes training examples that maximize influence improvement for the most underrepresented task in the selected data. 
It fundamentally differs from LESS, which scores each training example independently and selects the top-ranked ones.

\begin{table*}[t]
  \centering
  \resizebox{\linewidth}{!}{
  \begin{tabular}{@{}cccccccccc@{}}
    \toprule
    \multirow{2}{*}{\textbf{Budget}} &
      \multirow{2}{*}{\textbf{Method}} &
      \multicolumn{2}{c}{\textbf{Coding}} &
      \textbf{Logic} &
      \textbf{Knowledge} &
      \multicolumn{2}{c}{\textbf{Math}} &
      \textbf{Ins-Following} &
      \multirow{2}{*}{\textbf{Macro Avg}} \\ \cmidrule(lr){3-9}
     &
       &
      HumanEval &
      MBPP &
      BBH &
      MMLU &
      GSM-Plus &
      MATH &
      IFEval &
       \\ \midrule
    \multirow{6}{*}{5\%} &
      Random &
      43.5 &
      48.9 &
      {\ul 64.8} &
      64.9 &
      41.5 &
      22.5 &
      18.1 &
      43.4 \\
     &
      Task-max (LESS) &
      43.9 &
      50.7 &
      62.7 &
      \textbf{65.1} &
      \textbf{42.5} &
      22.6 &
      {\ul 19.7} &
      {\ul 43.9} \\
     &
      RDS &
      \textbf{45.6} &
      \textbf{52.7} &
      62.2 &
      65.0 &
      34.5 &
      17.2 &
      15.5 &
      41.8 \\
     &
      S2L &
      40.4 &
      49.0 &
      64.3 &
      65.0 &
      41.8 &
      \textbf{23.5} &
      16.5 &
      42.9 \\
     &
      DEITA &
      43.9 &
      47.3 &
      \textbf{65.0} &
      \textbf{65.1} &
      41.9 &
      22.3 &
      18.1 &
      43.4 \\
     &
      \textbf{BIDS (ours)} &
      \textbf{45.6} &
      {\ul 51.0} &
      64.3 &
      64.9 &
      {\ul 42.1} &
      {\ul 22.9} &
      \textbf{21.4} &
      \textbf{44.6} \\ \midrule
    \multirow{6}{*}{10\%} &
      Random &
      47.8 &
      50.6 &
      65.0 &
      64.9 &
      43.9 &
      24.0 &
      17.8 &
      44.9 \\
     &
      Task-max (LESS) &
      44.7 &
      {\ul 51.3} &
      62.0 &
      64.7 &
      {\ul 44.6} &
      24.3 &
      {\ul 19.3} &
      44.4 \\
     &
      RDS &
      {\ul 50.0} &
      \textbf{54.7} &
      63.2 &
      64.6 &
      39.3 &
      22.4 &
      18.3 &
      44.6 \\
     &
      S2L &
      46.5 &
      50.7 &
      64.5 &
      \textbf{65.0} &
      43.3 &
      22.7 &
      18.9 &
      44.5 \\
     &
      DEITA &
      \textbf{51.8} &
      49.9 &
      \textbf{65.6} &
      \textbf{65.0} &
      43.1 &
      {\ul 24.5} &
      17.5 &
      {\ul 45.3} \\
     &
      \textbf{BIDS (ours)} &
      48.2 &
      50.4 &
      {\ul 65.1} &
      64.9 &
      \textbf{45.1} &
      \textbf{25.1} &
      \textbf{23.4} &
      \textbf{46.0} \\ \midrule
    \multirow{7}{*}{15\%} &
      Random &
      48.7 &
      {\ul 51.9} &
      65.2 &
      \textbf{65.1} &
      {\ul 45.6} &
      {\ul 25.0} &
      18.8 &
      {\ul 45.7} \\
     &
      Task-max (LESS) &
      46.5 &
      51.0 &
      63.2 &
      64.6 &
      44.9 &
      24.9 &
      {\ul 21.2} &
      45.2 \\
     &
      RDS &
      {\ul 50.0} &
      \textbf{53.9} &
      63.7 &
      64.5 &
      41.1 &
      23.5 &
      18.1 &
      45.0 \\
     &
      S2L &
      \textbf{51.8} &
      50.7 &
      {\ul 65.4} &
      {\ul 64.8} &
      44.7 &
      24.2 &
      16.9 &
      45.5 \\
     &
      DEITA &
      {\ul 50.0} &
      50.7 &
      \textbf{65.8} &
      64.7 &
      45.0 &
      23.3 &
      16.7 &
      45.2 \\
     &
      \textbf{BIDS (ours)} &
      49.1 &
      50.7 &
      63.7 &
      64.6 &
      \textbf{45.8} &
      \textbf{26.2} &
      \textbf{22.6} &
      \textbf{46.1} \\ \cmidrule(l){2-10} 
     &
      \textbf{BIDS (ours; epochs=4)} &
      50.0 &
      53.0 &
      64.4 &
      \textbf{64.7} &
      47.0 &
      26.9 &
      \textbf{23.4} &
      \textbf{47.1} \\ \midrule
    \multirow{2}{*}{100\%} &
      Full (epochs=1) &
      \textbf{52.6} &
      53.6 &
      \textbf{65.5} &
      64.1 &
      47.2 &
      27.9 &
      17.5 &
      46.9 \\
     &
      Full (epochs=4) &
      48.2 &
      \textbf{54.4} &
      59.2 &
      63.1 &
      \textbf{51.5} &
      \textbf{32.3} &
      17.9 &
      46.7 \\ \bottomrule
    \end{tabular}%
  }
  \caption{
  Comparison between \name and various influence-and non-influence-based selection algorithms, with \textbf{Llama-3-8B} as the base model.
  The task-specific or macro-average performance is \textbf{bolded} if it ranks first under the same budget, and \underline{underlined} if it ranks second.
  ``\name (epochs=4)'' is compared with 100\% full training.
  }
  \label{tab:main_results-llama_select}
\end{table*}

\begin{table}[htb]
  \centering
  \resizebox{\linewidth}{!}{
  \begin{tabular}{@{}ccccc@{}}
    \toprule
    & \textbf{5\%}  & \textbf{10\%} & \textbf{15\%}      & \textbf{100\%}             \\ 
    \cmidrule(r){1-1} \cmidrule(l){2-5}
    Random          & 40.2          & 41.0          & 41.4               & \multirow{6}{*}{43.0/43.2\text{\scriptsize{(4)}}} \\
    Task-max (LESS) & 41.0          & 41.4          & 41.8               &                            \\
    RDS             & 38.5          & 39.6          & 41.0               &                            \\
    S2L             & 39.5          & 41.0          & 40.7               &                            \\
    DEITA           & 40.5          & 40.2          & 40.8               &                            \\
    \textbf{BIDS}   & \textbf{41.0} & \textbf{42.5} & \textbf{42.6}/\textbf{43.3}\text{\scriptsize{(4)}} &                            \\ \bottomrule
    \end{tabular}%
  }
  \caption{
  Macro-average with \textbf{Mistral-7B-v0.3} as the base model. Subscript \text{\scriptsize{(4)}} denotes the result of training for 4 epochs. 
  \name shows generalizability to different models.
  See Appendix \ref{append:6_mistral} for task-specific results.
  }
  \label{tab:main_results-mistral_select}
\end{table}

\section{Experiments}\label{sec:experiments}

\subsection{Experimental Setups}
We follow the setup in \S\ref{sec:analysis}, including the same LLMs, datasets, and influence estimation implementation (Table \ref{tab:main_results-llama_select}, \ref{tab:main_results-influence}). To further validate the generalizability of \name, we also experiment with base models from other model families (Table \ref{tab:main_results-mistral_select}).

\noindent\textbf{Baselines.} 
In addition to (1) the \textbf{random baseline} and (2) the original influence-based \textbf{task-wise max} algorithm used by LESS (\S\ref{sec:analysis}), 
we compare with three strong non-influence-based methods to further demonstrate the necessity of balanced data selection: 
(3) \textbf{Representation-based Data Selection (RDS; \citealp{zhang2018unreasonable, hanawa2020evaluation, xia2024less})} is a \textbf{targeted} selection method that also requires validation data representing the targeted capabilities. 
It selects training examples that are most similar to validation data based on the model's hidden representations. 
(4) \textbf{S2L (\citealp{yang2024smalltolarge})} and (5) \textbf{DEITA (\citealp{liu2024what})} are two state-of-the-art, \textbf{general} selection methods. 
They aim to select high-quality and diverse data without knowing the target tasks beforehand. 
Additionally, we compare to three other influence-based intuitive variants (6) \textbf{instance-wise max}, (7) \textbf{sum}, and (8) \textbf{temperature-based sampling} that are applicable to the Attribution Matrix in Table \ref{tab:main_results-influence}.
Appendix \ref{append:3_baselines} includes more baseline details.

\subsection{Results}\label{sec:main_results}

\paragraph{Comparison under the same budget.} 
As shown in Table \ref{tab:main_results-llama_select} and \ref{tab:main_results-mistral_select}, 
across different budget levels and models, 
\name consistently outperforms all baselines in terms of the macro-average across seven benchmarks.
On specific tasks, \name is frequently among the strongest, ranking either first or second on 4/7 benchmarks on average.
These results demonstrate that \name indeed achieves strong and balanced performance across tasks.
Finally, results in Table \ref{tab:main_results-influence} show the uniformly poor performance of influence-based variants other than \name, suggesting the necessity of going beyond simple Top-K selection, and taking normalization and cross-sample interactions into consideration.

Notably, all non-influence-based methods (RDS, S2L, DEITA) consistently underperform the random baseline across all budgets. 
Both RDS and DEITA are significantly biased towards the coding tasks under the 10\% and 15\% budgets, at the cost of serious performance drop on others, especially math and instruction-following. In contrast, S2L shows a shifting trend in its performance bias, from MATH under the 5\% to HumanEval under the 15\%.
These observations confirm the value of \name, as existing works on both targeted and general instruction data selection all fail to generalize to the challenging setup of learning multiple diverse capabilities simultaneously.

\begin{table}[htb]
  \centering
  \resizebox{\linewidth}{!}{
  \begin{tabular}{@{}ccccccc@{}}
    \toprule
    & \multicolumn{3}{c}{\textbf{Llama-3-8B}}       & \multicolumn{3}{c}{\textbf{Mistral-7B-v0.3}}  \\ 
    \cmidrule(lr){2-4} \cmidrule(lr){5-7}
    & 5\%  & 10\% & 15\% & 5\%  & 10\% & 15\% \\ 
    \cmidrule(r){1-1} \cmidrule(l){2-7} 
    Task-max     & 43.9 & 44.4 & 45.2 & 41.0 & 41.4 & 41.8 \\
    Instance-max & 44.2 & 45.0 & 45.0 & 39.7 & 41.1 & 42.0 \\
    Sum          & 44.2 & 44.6 & 44.9 & 39.9 & 40.2 & 41.2 \\
    Temperature  & 43.6 & 44.6 & 45.0 & 39.6 & 40.3 & 41.4 \\
    \textbf{BIDS} & \textbf{44.6} & \textbf{46.0} & \textbf{46.1} & \textbf{41.0} & \textbf{42.5} & \textbf{42.6} \\ \bottomrule
    \end{tabular}%
  }
  \caption{
  Macro-average of influence-based algorithms with \textbf{Llama-3-8B}
  and \textbf{Mistral-7B-v0.3}. 
  See Appendix \ref{append:5_llama}
  and \ref{append:6_mistral}
  for task-specific results.
  }
  \label{tab:main_results-influence}
\end{table}

\begin{figure*}[t] %
    \centering
    \begin{subfigure}{0.3\textwidth}        
        \input{figures/THI_unnorm_col_p0.1}
    \end{subfigure}
    \hfill %
    \begin{subfigure}{0.3\textwidth}
        \input{figures/THI_norm_col_p0.1}
    \end{subfigure}
    \hfill %
    \begin{subfigure}{0.3\textwidth}
        \input{figures/THI_BIDS_p0.1}
    \end{subfigure}
    \caption{THI under the 10\% budget with \textbf{Llama-3-8B}. %
    Both \textbf{$- \texttt{Iter}$} and \name have more balanced task frequencies.
    }
    \label{fig:compare_THI}
\end{figure*}

\begin{figure*}[t] %
    \centering
    \begin{subfigure}{0.3\textwidth}
        
        \input{figures/AID_unnorm_col_p0.1}
    \end{subfigure}
    \hfill %
    \begin{subfigure}{0.3\textwidth}

        \input{figures/AID_norm_col_p0.1}
    \end{subfigure}
    \hfill %
    \begin{subfigure}{0.3\textwidth}

        \input{figures/AID_BIDS_p0.1}
    \end{subfigure}
    \caption{
    Normalized AID under the 10\% budget with \textbf{Llama-3-8B}. %
    Disparity in AID 
    gradually diminishes, with both decreasing maximums and increasing minimums.
    Appendix \ref{append:6_mistral} shows similar trends with Mistral-7B-v0.3.
    }
    \label{fig:compare_AID}
\end{figure*}

\paragraph{\name outperforms full-dataset training.}
As shown in the last three rows of Table \ref{tab:main_results-llama_select} and the last two columns of Table \ref{tab:main_results-mistral_select}, finetuning on a 15\% subset selected by \name over four epochs consistently outperforms training on the full dataset. 
Task-specific results further reveal that \name achieves better performance by maintaining balanced and strong performance across six reasoning-related tasks while significantly improving general instruction-following. 
These results demonstrate that \name not only excels in selecting influential and balanced data, but also that full-dataset training may not always be optimal for LLMs to learn multiple diverse capabilities,
highlighting the efficiency potential of selective instruction finetuning.

\section{Ablation}\label{sec:ablation}

\begin{figure*}[htbp]
  \centering
  \begin{minipage}{.45\textwidth}
    \input{figures/corr_length_col}
  \end{minipage}
  \hspace{0.02\textwidth} %
  \begin{minipage}{.45\textwidth}
    \input{figures/corr_length_task}
  \end{minipage}
\end{figure*}

This section presents ablation studies of the two key components of BIDS, 
and analyzes their effects on the composition of selected data.

\paragraph{Ablation.}
We compare \name with the \textbf{$- \texttt{Iter}$} baseline to ablate iterative selection, and with \textbf{$- (\texttt{Norm} + \texttt{Iter})$} to further ablate normalization (i.e., the \textbf{instance-wise max} algorithm in Appendix \ref{append:3_baselines}).
Table \ref{tab:sep_contrib} shows that both the iterative selection and normalization  consistently improve the overall performance across all budgets.

\begin{table}[htb]
  \centering
  \resizebox{\linewidth}{!}{
  \begin{tabular}{@{}ccccccc@{}}
    \toprule
    & \multicolumn{3}{c}{\textbf{Llama-3-8B}}       & \multicolumn{3}{c}{\textbf{Mistral-7B-v0.3}}  \\ 
    \cmidrule(lr){2-4} \cmidrule(lr){5-7}
    & 5\%  & 10\% & 15\% & 5\%  & 10\% & 15\% \\ 
    \cmidrule(r){1-1} \cmidrule(l){2-7}
    \textbf{BIDS} & \textbf{44.6} & \textbf{46.0} & \textbf{46.1} & \textbf{41.0} & \textbf{42.5} & \textbf{42.6} \\
    $- \texttt{Iter}$                   & 44.3 & 45.6 & 45.7 & 41.0 & 41.6 & 42.2 \\
    $- (\texttt{Norm} + \texttt{Iter})$ & 44.2 & 45.0 & 45.0 & 39.7 & 41.1 & 42.0 \\ \bottomrule
    \end{tabular}%
  }
  \caption{
  Respective contributions of the two components of \name. 
  Results with both models uniformly show a decreasing trend as more technical components are ablated, substantiating their positive contributions.
  }
  \label{tab:sep_contrib}
\end{table}

\paragraph{Normalization balances THI.}
We then proceed to explore how the two components affect the influence distribution of selected data, and whether such effects can provide insights into how \name advances balanced learning of diverse capabilities.
We replace task-wise average influence with instance-wise influence in THI calculation,
since the three algorithms we are comparing are all built upon the instance-wise max approach. 
Comparing \ref{fig:THI_unnorm_col_p0.1} with \ref{fig:THI_norm_col_p0.1}, 
 after normalization the task frequency distribution becomes more balanced. Frequencies for MMLU, GSM-Plus, MATH and IFEval all increase by a great extent, while those for BBH and coding tasks decrease. 
\ref{fig:THI_norm_col_p0.1} to \ref{fig:THI_BIDS_p0.1} shows a few mild adjustments made by the iterative selection.
This is intriguing when compared to the task-specific results in Table \ref{tab:main_results-llama_select}, where \name actually shows superior performance in tasks with both increased (HumanEval, IFEval) and decreased (GSM-Plus, MATH) THI frequencies.
This suggests that a balanced selection may improve data efficiency not only by allocating more budget for underrepresented capabilities, but also reducing the redundancy in over-represented ones.

\paragraph{Better performance comes with smaller influence discrepancies.}
We report influence values after instance-level normalization
for better comparisons.
From~\ref{fig:AID_unnorm_col_p0.1} to~\ref{fig:AID_norm_col_p0.1} to~\ref{fig:AID_BIDS_p0.1}, we observe a progressive reduction in the disparity of average influence across tasks, which leads to the following two interesting observations:
(1) \textbf{The maximums of AID decrease.} Despite generally lower influence scores across the seven tasks, the performance of \name improves consistently. This reveals a limitation of the first-order linearity assumption by the influence estimation method: 
simply selecting high-influence points using a 
greedy
algorithm increases the average influence distribution on almost all tasks, but their effectiveness doesn't linearly add up, thus not necessarily improving task-level or overall performance.
(2) \textbf{The minimums of AID increase, especially for validation instances with exceptionally low influence}, such as HumanEval and MBPP. This again suggests the effectiveness of \name' key motivation: improving the model's overall performance by enhancing the capabilities that are most underrepresented.

\section{Completion Length as a Task-specific Factor for Spurious Influence}\label{sec:factor_analysis}
This analysis aims to explain the counterintuitive phenomenon identified in \S\ref{sec:analysis}, where the performance of high-influence tasks actually degrades (e.g., MMLU).
We focus on the \textbf{completion length of target tasks}, %
which 
has been shown important for gradient-based influence and data selection \citep{wang2024a, liu2024what, xia2024less}.
Ideally, if influence has a salient correlation with a task-specific factor that is \textbf{not strongly correlated with downstream performance}, then this can be explained, as such spurious influence no longer translates into better performance.

Figures \ref{fig:corr_length_col} and \ref{fig:corr_length_task} show instance- and task-level correlations between influence and completion lengths of validation data. 
\textbf{On the task-level, there is a strong negative correlation ($R = -0.769$) for 6/7 tasks.} 
The shortest completion length of MMLU indeed translates into the highest average influence.
Only IFEval, the only task that does not have a single objective answer, contradicts this negative correlation,
possibly suggesting other influencing factors.
However,
\textbf{the instance-level correlation pattern becomes more intricate}, with a global correlation of only $-0.083$.
Even for tasks that show strong negative correlations in Figure \ref{fig:corr_length_task} such as MMLU, HumanEval, MBPP and GSM-Plus, there is no similar trend for instances inside the same task, not to mention 
IFEval, which even shows a slightly positive correlation. 

In summary,
completion lengths can partly explain the counterintuitive phenomenon, and should be considered for improving the fidelity of influence.
But it is still far from the whole picture, suggesting the potential impact of other task-specific factors such as sample style or complexity \citep{liu2024what}, which are harder to quantify and deserve more focused investigation in future work.

\section{Related Work}
\paragraph{Data Selection for Instruction Finetuning.}
Since \citet{zhou2024lima} showed 
the importance of high quality instruction data curation, 
many works have been exploring automatic data selection methods. 
Quality-guided selection defines the quality for each data point using heuristic indicators based on model embeddings or losses~\citep{yang2023decoding, ankner2025perplexed, zhang2025best}, ratings from strong evaluators
~\citep{chen2023alpagasus, parkar2024selectllm, pang2025improving}, or principled metrics based on
training dynamics~\citep{kang2024get, mekala-etal-2024-smaller, xia2024less, lin2025lead}. Diversity-guided methods usually cluster the training data based on informative representations~\citep{yang2024smalltolarge, zhang2025harnessing}, and take inspiration from traditional core-set selection approaches~\citep{das2023deft}.
Both have been proved effective for instruction finetuning of LLMs~\citep{bukharin2023data, liu2024what, qin2025unleashing},
and \name takes both quality and diversity into account by applying an iterative selection algorithm to influence distributions.
Additionally, there is also a line of work that surveys how existing data selection techniques perform in compute-constrained scenarios~\citep{yin2025computeconstrained}, under different experimental configurations~\citep{diddee-ippolito-2025-chasing}, and for large-scale datasets~\citep{ivison2025large}.

\paragraph{Influence Estimation.} 
Influence estimation 
is generally classified into gradient-based and retraining-based approaches~\citep{hammoudeh2024training, ko2024mirrored, deng:hal-05230469}. Gradient-based influence estimation focuses on the gradient trace of each training point, and assesses the gradient alignment between training and validation examples~\citep{koh2017understanding, pruthi2020estimating, pan2025daunce}. Retraining-based estimation trains a large number of models on different subsets, and then inspects how their performance changes when a training example is added to these subsets~\citep{ghorbani2019data, ilyas2022datamodels, park2023trak, hu2025grass}. Recently both lines of work have been extended to LLM-scale applications, covering both pretraining~\citep{engstrom2024dsdm, yu2024mates, choe2024your, chang2025scalable, yu2025data, wang2025capturing} and instruction tuning~\citep{xia2024less, liu2024less, pan-etal-2024-g}.
There is also a popular line of work aimed at designing efficient proxies for expensive gradient-based or retraining-based estimation by taking inspiration from in-context probing~\citep{jiao-etal-2025-feasibility} and uncertainty estimation~\citep{pan2025daunce}.

\paragraph{Gradient-induced Spurious Influence.} Although to the best of our knowledge, our work is the first to systematically identify and address the effect of spurious influence on LLM training data selection, a few past works have also offered theoretical or empirical discussions on gradient-induced spurious influence in different machine learning applications. 
\citet{xia2024less} attribute spurious influence to the  representation of example-wise gradients, which is most often made tractable by taking an average over the gradients of all tokens inside each example. 
This likely leads to the undesirable consequence that shorter training examples (i.e., with fewer tokens) 
have inherently higher influence values when the influence is computed by gradient dot product,
because the norm of averaged token-wise gradients tends to rapidly decay with the length of example.
In this work, we follow \citet{xia2024less} by adopting cosine similarity-based (i.e., dot product with length normalization) influence formulation, which should ideally remove the length bias. However, our empirical results in \S\ref{sec:factor_analysis} suggest that the correlation between influence values and example length still remains for normalized gradients.
One possible reason is that taking the average over token-wise gradients may incur systematic bias in the direction of the resulting gradient as well. 

Apart from the application of LLMs in natural language generation tasks, past works have also identified spurious influence for BERT~\citep{devlin2019bert} in traditional NLP tasks, and for diffusion models in image generation. \citet{yeh2022first} identify the bias of gradient norms across different layers of a BERT model, and attribute this to a ``cancellation effect'' in certain weight parameters. \citet{xie2024data} discover the bias of gradient norms over different timesteps in the generation process of diffusion models, where the loss function component from later timesteps is more likely to have a larger gradient norm. In light of these, we suggest in-depth future efforts into the possible factors that may systematically affect both the norm and direction of example gradients.

\section{Conclusion}
We introduce \name, an influence-based instruction tuning data selection algorithm specifically designed for learning multiple  capabilities. Motivated by the observation of an inherent bias in influence across different tasks, \name first applies column-wise normalization to the Attribution Matrix, and then an iterative selection algorithm favoring underrepresented tasks. \name consistently outperforms various baselines as well as full-dataset training with much more balanced performance. Our analysis provides insights into the properties of an influential dataset with balanced capabilities.

\section*{Limitations}\label{limitations}
We discuss the limitations of our work below.

\begin{enumerate}
    \item \textbf{Limitation to influence-based approaches.} While our work focuses on the imbalance issue of influence-based data selection methods, results of non-influence-based methods (i.e., RDS, D2L, DEITA) in Table \ref{tab:main_results-llama_select} also show significant bias towards the two coding tasks, at the cost of severely degraded performance on others. These observations suggest the possibility that the imbalance of utility scores~\citep{yin2025computeconstrained} may generally exist for both influence- and non-influence-based data selection approaches. 
    In Appendix~\ref{append:7_RDS_BIDS}, we conduct a preliminary investigation by applying the technical components of \name to one of the non-influence-based methods, RDS, and find that the balancing techniques of \name work well for RDS, which provides a preliminary verification for our hypothesis above, and also shows the broad applicability of the underlying principles of \name.
    However, the focus of this paper
    on influence-based approaches
    limits in-depth investigations into the more general imbalance of utility scores for data selection under a multi-capability learning setup. We hope more analyses and technical solutions can be discussed in future work.

    \item \textbf{More factors behind the influence bias.} While our work centers on identifying the problem of imbalanced influence and proposing \name as a remedy, probing the factors behind the influence bias is also beneficial for a thorough understanding of the counterintuitive phenomenon identified in \S\ref{sec:analysis}. 
    In \S\ref{sec:factor_analysis}
    we provide a preliminary analysis on the completion length of validation data as a potential factor. Though it can partly account for the phenomenon above, it is still far from a complete explanation, which requires investigation into other task-specific factors that are much harder to quantify. We hope future work can shed light on this.
\end{enumerate}

\section*{Acknowledgments}\label{acknowledgements}
We thank Lifan Yuan and Qingzhi Chen for their insightful discussions on this project. We also sincerely appreciate the anonymous ARR reviewers and chairs for their helpful feedback, which greatly improved this work.

\bibliography{custom}

\clearpage
\appendix

\section{Appendix}

\subsection{Influence Estimation Pipeline of LESS}

We briefly introduce the influence estimation pipeline of LESS in this section. For more detailed motivation and step-by-step mathematical deduction, we suggest referring to~\citet{xia2024less}.

Assume a model $\gM_s$ which scores and selects data, and another model $\gM_t$ which is trained on the selected data. For a training dataset $\gD$ and validation dataset $\gV$,  LESS formulates the pairwise influence between each training example $\vt_i \in \gD$ and validation instance $\vv_j \in \gV$ with the following three steps.

\paragraph{Step 1: Warmup training with LoRA.}  
LESS first trains $\gM_s$ on a random subset $\gD_\text{warmup} \subset \gD$ for $N$ epochs using the parameter-efficient finetuning method LoRA~\citep{hu2021lora}, and checkpoints the model after each epoch to store LoRA parameters $\{\vtheta_t\}_{t=1}^N$.

\paragraph{Step 2: Gradient computation and projection.} 
For each checkpoint $\vtheta_t$ of LoRA-trained $\gM_s$, LESS computes the SGD gradient of validation instance $\vv_j$, and further uses random projection~\citep{Johnson1984ExtensionsOL, park2023trak} to project the gradient to a tractable lower dimension. The resulting projected gradient is denoted as 
$\grad{\vv_j}{\vtheta_t}$
. LESS also computes and projects the gradient of training example $\vt_i$, but uses the Adam gradient defined as follows:
$$\update{\vt_i}{\vtheta_{t}} \triangleq \frac{\vm^{t+1}}{\sqrt{\vv^{t+1} + \epsilon}}$$
where $\vm^{t+1}$ and $\vv^{t+1}$ are the first and second moments in the parameter update rule for Adam optimizer.

\paragraph{Step 3: Gradient matching and influence calculation.} Finally, LESS employs the following cosine-similarity-based approach to calculate the alignment between the gradient of each training and validation example, accumulated over all the warmup training epochs:
$$\infadam(\vt_i, \vv_j) \triangleq \sum_{t=1}^{N} \bar{\eta_{t}} \cos(\grad{\vv_j}{\vtheta_t}, \update{\vt_i}{\vtheta_{t}})$$
where $\bar{\eta_{t}}$ is the average learning rate in the $t$-th epoch.

\subsection{Details of Training and Evaluation Setups}\label{append:2_setup}

Based on the LESS pipeline described above, we further introduce the implementation details of the training and evaluation setups in this work. All the experiments are carried out on 2 H100 GPUs with 80 GB memories.

\paragraph{Training Details.} We basically follow the same set of hyperparameters as LESS when training both $\gM_s$ and $\gM_t$. Specifically, a batch size of 128 is used throughout all the training processes in this work, along with a learning rate scheduler with linear warm-up, cosine decay, and a peak learning rate of $2 \times 10^{-5}$.
For the influence estimation pipeline, we consistently conduct the warmup training of $\gM_s$ using four epochs and the full training set. For gradient computation and projection, we uniformly sample 50 validation instances from either the validation or the test split (when there is not a separate validation split) of each of the seven evaluation tasks, leading to a total of 350 validation instances. The projection dimension is set as 8192 for all the training and validation examples. For training $\gM_t$ on the selected data, we consistently train for two epochs if not otherwise specified. Notably, we report the average performance of two sets of randomly selected data for all the random baselines in this paper.

Both the warmup training for influence estimation and the training on selected data are carried out with LoRA~\citep{hu2021lora}. The configurations of LoRA adapters are kept the same throughout the experiments, with a rank of 128, an $\alpha$ value of 512, a dropout rate of 0.1, and LoRA matrices being applied to all the attention modules.

\paragraph{Evaluation Details.} We follow the evaluation convention of UltraInteract~\citep{yuan2024advancing} by using greedy decoding (i.e., temperature $=$ 0) for all the evaluation tasks except for IFEval, where we use temperature $=$ 0.7 and take the median result of three random seeds due to the high variability of this task.

\subsection{Details of Data Selection Baselines}\label{append:3_baselines}

In this section, we first specify the mathematical definitions of the four \textbf{influence-based} selection algorithms used in this work. They share the same framework of first assigning an overall utility score $s_{i}$ to each training example $t_{i}$, and then greedily selecting examples with the highest scores, or randomly sampling from the score distribution.

\begin{itemize}
    \item \textbf{Task-wise Max:} $s_{i} \triangleq \max_{k=1, \dots, m}\{\sum_{\vv_{j} \in \gV_k} \mA_{ij}\}$. It first computes the \textbf{average} influence over validation instances within the same task, 
    and then takes the \textbf{maximum} influence across different tasks as the utility score.

    \item \textbf{Instance-wise Max:} $s_{i} \triangleq \max_{j=1, \dots, |\gV|} \{\mA_{ij}\}$. It directly uses the \textbf{maximum} of influence values over all validation instances as the utility score. Notably, instance-wise max is equivalent to \textbf{$- (\texttt{Norm} + \texttt{Iter})$} in \S\ref{sec:ablation}.

    \item \textbf{Sum:} $s_{i} \triangleq \sum_{j=1}^{|\gV|} \mA_{ij}$. It directly uses the \textbf{sum} of influence values over all validation instances as the utility score.

    \item \textbf{Temperature-based Sampling:} $s_{i}$ follows the same definition as \textbf{Task-wise Max}. 
    However, instead of greedy selection, this algorithm transforms these utility scores into a probability distribution using temperature-scaled softmax \citep{ficler-goldberg-2017-controlling, zhang2024edt}, and samples training data from this distribution without replacement.
    Compared with the deterministic greedy selection, this sampling-based approach increases the proportion of selected data that show preference for low-influence tasks, thus balancing the task distribution. 
\end{itemize}

We then summarize the mechanisms and provide the implementation details of the remaining three \textbf{non-influence-based} selection methods. 

\begin{itemize}
    \item \textbf{RDS \citep{zhang2018unreasonable, hanawa2020evaluation, xia2024less}.} We follow the RDS implementation in \citet{xia2024less}, which computes the cosine similarity between training and validation examples based on the final layer representation of the last token in each example sequence. 
    Training examples with the highest similarity to any one of the validation examples are selected. In order to ensure fair comparison, we use the same model that computes gradient features in \name to extract the final layer representations for RDS.

    \item \textbf{S2L \citep{yang2024smalltolarge}} first collects the training loss trajectory for each training example using a small reference model that belongs to the same family as the final model, and then applies K-means clustering to these trajectories. Finally, an equal number of examples are sampled from each cluster to train the final model, ensuring the balance among clusters of different sizes.
    Specifically, we use Llama-3.2-1B%
    \footnote{\url{https://huggingface.co/meta-llama/Llama-3.2-1B}}%
    as the reference model, which belongs to the Llama-3 family and is much larger than the Pythia-70M \citep{biderman2023pythia} used in \citet{yang2024smalltolarge}, thus improving the fidelity of S2L's performance. The remaining hyperparameters follow the configurations in \citet{yang2024smalltolarge}. 
    We note that although S2L performs selection with a small model, its total computational cost is similar to LESS, as the loss for each example needs to be computed more frequently than the gradient along the training trajectory.

    \item \textbf{DEITA \citep{liu2024what}} first uses ChatGPT \citep{achiam2023gpt} to curate training data for training a complexity scorer and a quality scorer. Then with the trained scorers, it assigns a complexity score and a quality score to each training example and sorts the training dataset by their product. Finally, it applies an iterative selection algorithm to the sorted dataset based on the embedding of each training example, with the aim of promoting diversity. Specifically, we use the 7B scorer models%
    \footnote{\url{https://huggingface.co/hkust-nlp/deita-complexity-scorer}}%
    \footnote{\url{https://huggingface.co/hkust-nlp/deita-quality-scorer}}%
    publicly released by the DEITA team, and obtain example-wise embeddings using the final-layer hidden states of a pretrained Llama-3-8B to align with the final evaluation.
\end{itemize}

\subsection{Effect of Normal Standardization on the Attribution Matrix}\label{append:4_norm_effect}

In \S\ref{sec:bids} we introduce the instance-level normalization technique of \name. One potential issue with this normal standardization approach is that it may not work sufficiently well when the distribution of unnormalized influence scores differs much from an approximate normal distribution. 
In this section we aim to justify the application of normal standardization to the Attribution Matrix (AM). Specifically, we randomly select five validation instances (i.e., five columns in the AM) from each task, and compare their empirical distributions after normalization with a standard normal distribution. The results show that almost all of the sampled columns approximate a standard normal distribution after the instance-level normalization, which justifies the use of normal standardization as the normalization method in \name.

\begin{figure*}[ht]
    \centering
    \includegraphics[width=0.95\linewidth]{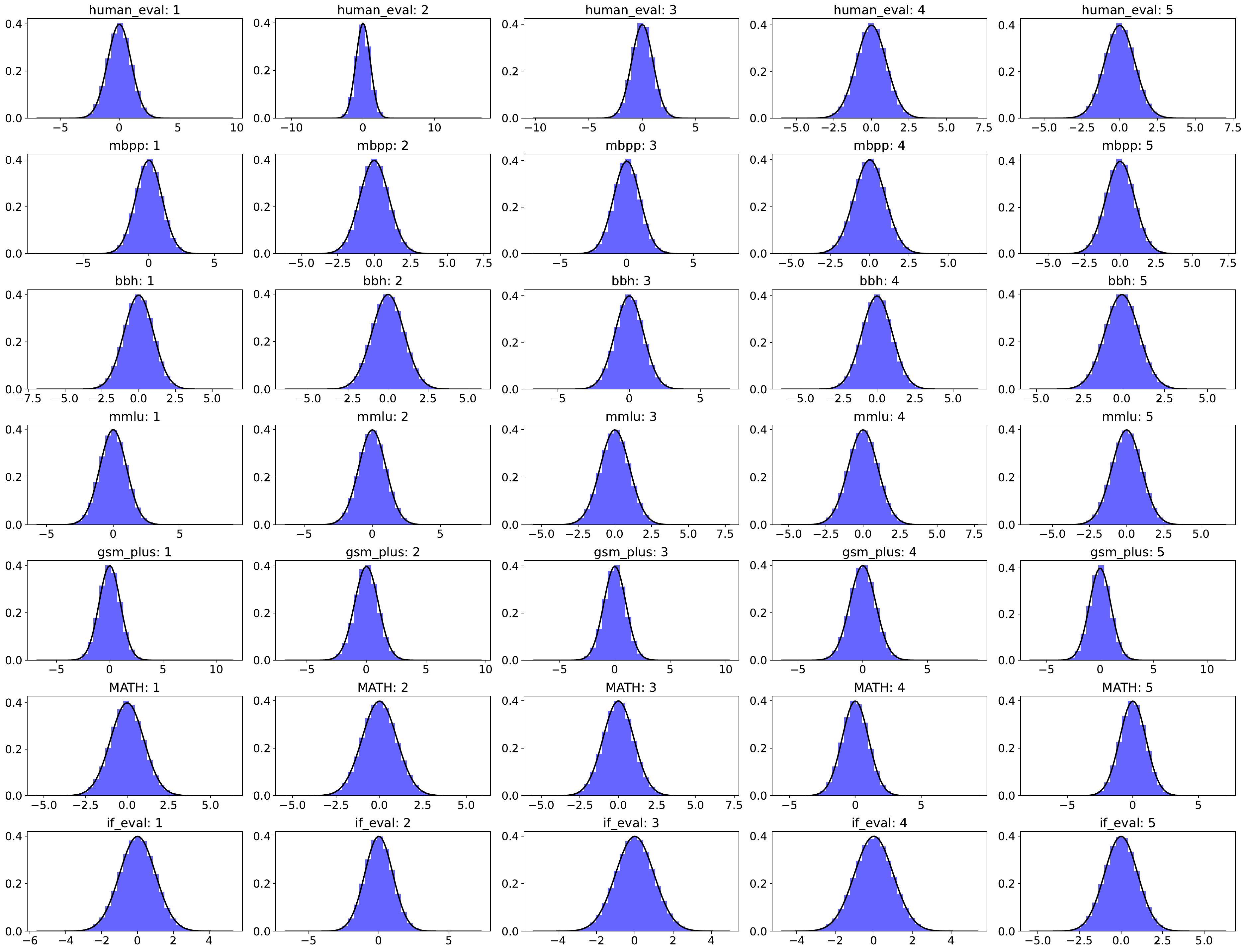}
    \caption{The effect of normal standardization. Five AM columns are sampled for each task. Most of the columns in the AM indeed approximate a standard normal distribution after normal standardization.}
    \label{fig:norm_effect}
\end{figure*}

\subsection{Complete Results with Llama-3-8B}\label{append:5_llama}

In Table \ref{tab:append_results-llama_select}, we present the complete task-specific results for Table \ref{tab:main_results-influence} and \ref{tab:sep_contrib} in the main text. 
Across all three budgets, apart from the consistently strongest macro-average, \name also ranks either first or second on 4/7, 5/7, 4/7 benchmarks respectively, showing its stronger and more balanced performance than both simple influence-based variants and ablated baselines. 
This further highlights the necessity of going beyond the naive greedy Top-K selection or temperature-based random sampling, and taking both normalization and interactions among training examples into consideration, in order to tackle spurious or biased cross-task influence.

\begin{table*}[ht]
  \centering
  \resizebox{\linewidth}{!}{
  \begin{tabular}{@{}cccccccccc@{}}
    \toprule
    \multirow{2}{*}{\textbf{Budget}} &
      \multirow{2}{*}{\textbf{Method}} &
      \multicolumn{2}{c}{\textbf{Coding}} &
      \textbf{Logic} &
      \textbf{Knowledge} &
      \multicolumn{2}{c}{\textbf{Math}} &
      \textbf{Ins-Following} &
      \multirow{2}{*}{\textbf{Macro Avg}} \\ \cmidrule(lr){3-9}
     &
       &
      HumanEval &
      MBPP &
      BBH &
      MMLU &
      GSM-Plus &
      MATH &
      IFEval &
       \\ \midrule
    \multirow{6}{*}{5\%} &
      Task-max (LESS) &
      43.9 &
      50.7 &
      62.7 &
      \textbf{65.1} &
      {\ul 42.5} &
      22.6 &
      19.7 &
      43.9 \\
     &
      Sum &
      \textbf{45.6} &
      51.9 &
      63.6 &
      64.8 &
      42.4 &
      21.3 &
      20.1 &
      44.2 \\
     &
      Temperature &
      44.7 &
      47.6 &
      {\ul 64.2} &
      {\ul 65.0} &
      42.4 &
      \textbf{23.7} &
      17.7 &
      43.6 \\
     &
      \textbf{BIDS (ours)} &
      \textbf{45.6} &
      51.0 &
      \textbf{64.3} &
      64.9 &
      42.1 &
      {\ul 22.9} &
      \textbf{21.4} &
      \textbf{44.6} \\
     &
      $- \texttt{Iter}$ &
      \textbf{45.6} &
      \textbf{52.1} &
      62.5 &
      64.8 &
      {\ul 42.5} &
      22.5 &
      20.1 &
      {\ul 44.3} \\
     &
      $- (\texttt{Norm} + \texttt{Iter})$ &
      43.9 &
      \textbf{52.1} &
      63.2 &
      {\ul 65.0} &
      \textbf{42.6} &
      22.3 &
      {\ul 20.6} &
      44.2 \\ \midrule
    \multirow{6}{*}{10\%} &
      Task-max (LESS) &
      44.7 &
      {\ul 51.3} &
      62.0 &
      64.7 &
      44.6 &
      24.3 &
      19.3 &
      44.4 \\
     &
      Sum &
      45.6 &
      \textbf{51.6} &
      61.6 &
      64.6 &
      43.8 &
      23.7 &
      21.0 &
      44.6 \\
     &
      Temperature &
      44.7 &
      50.4 &
      64.4 &
      64.9 &
      44.9 &
      24.2 &
      18.9 &
      44.6 \\
     &
      \textbf{BIDS (ours)} &
      \textbf{48.2} &
      50.4 &
      \textbf{65.1} &
      64.9 &
      {\ul 45.1} &
      {\ul 25.1} &
      \textbf{23.4} &
      \textbf{46.0} \\
     &
      $- \texttt{Iter}$ &
      {\ul 47.4} &
      48.4 &
      {\ul 64.6} &
      \textbf{65.1} &
      \textbf{45.4} &
      \textbf{25.2} &
      {\ul 23.0} &
      {\ul 45.6} \\
     &
      $- (\texttt{Norm} + \texttt{Iter})$ &
      46.5 &
      47.3 &
      {\ul 64.6} &
      {\ul 65.0} &
      44.1 &
      24.7 &
      22.8 &
      45.0 \\ \midrule
    \multirow{6}{*}{15\%} &
      Task-max (LESS) &
      46.5 &
      {\ul 51.0} &
      63.2 &
      64.6 &
      44.9 &
      24.9 &
      {\ul 21.2} &
      45.2 \\
     &
      Sum &
      {\ul 48.2} &
      {\ul 51.0} &
      62.6 &
      64.6 &
      44.8 &
      24.0 &
      19.3 &
      44.9 \\
     &
      Temperature &
      45.6 &
      \textbf{52.1} &
      \textbf{65.1} &
      64.7 &
      45.5 &
      23.6 &
      18.7 &
      45.0 \\
     &
      \textbf{BIDS (ours)} &
      \textbf{49.1} &
      50.7 &
      63.7 &
      64.6 &
      \textbf{45.8} &
      \textbf{26.2} &
      \textbf{22.6} &
      \textbf{46.1} \\
     &
      $- \texttt{Iter}$ &
      47.4 &
      50.1 &
      {\ul 64.9} &
      \textbf{65.0} &
      45.6 &
      {\ul 26.0} &
      20.8 &
      {\ul 45.7} \\
     &
      $- (\texttt{Norm} + \texttt{Iter})$ &
      47.4 &
      48.1 &
      63.2 &
      \textbf{65.0} &
      \textbf{45.8} &
      25.1 &
      20.3 &
      45.0 \\ \bottomrule
    \end{tabular}%
  }
  \caption{
  Complete results for Table \ref{tab:main_results-influence} and \ref{tab:sep_contrib} in the main text, with \textbf{Llama-3-8B} as the base model. 
  Notably, \textbf{$- (\texttt{Norm} + \texttt{Iter})$} (\S\ref{sec:ablation}; Table \ref{tab:sep_contrib}) is equivalent to the \textbf{instance-wise max} algorithm (Appendix \ref{append:3_baselines}; Table \ref{tab:main_results-influence}).
  }
  \label{tab:append_results-llama_select}
\end{table*}

\subsection{Complete Results with Mistral-7B-v0.3}\label{append:6_mistral}

To further validate the generalizability of \name, we compare with other baseline data selection algorithms using Mistral-7B-v0.3 as the backbone for both selection and training. The complete task-specific results are presented in Table \ref{tab:append_results-mistral_select-non_influence} and \ref{tab:append_results-mistral_select-influence}. The two algorithms compared here, \textbf{$- (\texttt{Norm} + \texttt{Iter})$} and \textbf{$- \texttt{Iter}$}, follow the same definition in 
\S\ref{sec:ablation}. 
And the random baseline is also the average result of two different random seeds.

\begin{table*}[!htb]
  \centering
  \resizebox{\linewidth}{!}{
  \begin{tabular}{@{}cccccccccc@{}}
    \toprule
    \multirow{2}{*}{\textbf{Budget}} &
      \multirow{2}{*}{\textbf{Method}} &
      \multicolumn{2}{c}{\textbf{Coding}} &
      \textbf{Logic} &
      \textbf{Knowledge} &
      \multicolumn{2}{c}{\textbf{Math}} &
      \textbf{Ins-Following} &
      \multirow{2}{*}{\textbf{Macro Avg}} \\ \cmidrule(lr){3-9}
     &
       &
      HumanEval &
      MBPP &
      BBH &
      MMLU &
      GSM-Plus &
      MATH &
      IFEval &
       \\ \midrule
    \multirow{6}{*}{5\%} &
      Random &
      36.8 &
      44.3 &
      {\ul 59.5} &
      61.7 &
      37.0 &
      \textbf{19.9} &
      22.2 &
      40.2 \\
     &
      Task-max (LESS) &
      36.8 &
      \textbf{45.6} &
      \textbf{60.1} &
      \textbf{62.3} &
      \textbf{38.6} &
      19.5 &
      {\ul 23.8} &
      41.0 \\
     &
      RDS &
      {\ul 38.6} &
      45.3 &
      58.2 &
      61.9 &
      33.3 &
      15.4 &
      16.9 &
      38.5 \\
     &
      S2L &
      36.8 &
      44.4 &
      58.9 &
      61.9 &
      36.5 &
      16.6 &
      21.2 &
      39.5 \\
     &
      DEITA &
      \textbf{43.0} &
      \textbf{45.6} &
      {\ul 59.5} &
      {\ul 62.0} &
      36.1 &
      17.2 &
      19.8 &
      40.5 \\
     &
      \textbf{BIDS (ours)} &
      37.7 &
      44.4 &
      {\ul 59.5} &
      61.8 &
      {\ul 38.0} &
      {\ul 19.8} &
      \textbf{26.1} &
      \textbf{41.0} \\ \midrule
    \multirow{6}{*}{10\%} &
      Random &
      37.7 &
      44.8 &
      59.8 &
      61.8 &
      {\ul 40.0} &
      {\ul 21.2} &
      22.0 &
      41.0 \\
     &
      Task-max (LESS) &
      39.5 &
      43.8 &
      {\ul 60.3} &
      \textbf{62.0} &
      39.7 &
      20.0 &
      {\ul 24.6} &
      41.4 \\
     &
      RDS &
      \textbf{43.9} &
      {\ul 45.0} &
      57.8 &
      61.7 &
      35.6 &
      14.9 &
      18.1 &
      39.6 \\
     &
      S2L &
      36.8 &
      44.7 &
      59.9 &
      61.7 &
      39.2 &
      \textbf{22.0} &
      22.4 &
      41.0 \\
     &
      DEITA &
      36.8 &
      {\ul 45.0} &
      59.1 &
      \textbf{62.0} &
      39.3 &
      18.6 &
      20.4 &
      40.2 \\
     &
      \textbf{BIDS (ours)} &
      {\ul 40.4} &
      \textbf{46.1} &
      \textbf{60.5} &
      61.7 &
      \textbf{40.5} &
      21.0 &
      \textbf{27.1} &
      \textbf{42.5} \\ \midrule
    \multirow{7}{*}{15\%} &
      Random &
      37.7 &
      45.9 &
      \textbf{59.6} &
      \textbf{61.9} &
      41.0 &
      \textbf{21.7} &
      21.8 &
      41.4 \\
     &
      Task-max (LESS) &
      {\ul 40.4} &
      44.1 &
      \textbf{59.6} &
      61.3 &
      \textbf{41.9} &
      21.5 &
      {\ul 24.0} &
      41.8 \\
     &
      RDS &
      \textbf{48.2} &
      \textbf{48.1} &
      58.4 &
      {\ul 61.7} &
      36.6 &
      16.4 &
      17.9 &
      41.0 \\
     &
      S2L &
      34.2 &
      46.7 &
      58.8 &
      {\ul 61.7} &
      40.3 &
      20.9 &
      22.6 &
      40.7 \\
     &
      DEITA &
      37.7 &
      46.1 &
      58.5 &
      61.5 &
      40.1 &
      20.3 &
      21.6 &
      40.8 \\
     &
      \textbf{BIDS (ours)} &
      39.5 &
      {\ul 47.0} &
      59.5 &
      {\ul 61.7} &
      {\ul 41.8} &
      \textbf{21.7} &
      \textbf{27.1} &
      \textbf{42.6} \\ \cmidrule(l){2-10} 
     &
      \textbf{BIDS (ours; epochs=4)} &
      40.4 &
      47.0 &
      \textbf{58.9} &
      \textbf{61.1} &
      44.1 &
      23.5 &
      \textbf{28.1} &
      \textbf{43.3} \\ \midrule
    \multirow{2}{*}{100\%} &
      Full (epochs=1) &
      40.4 &
      \textbf{49.6} &
      58.8 &
      60.8 &
      45.2 &
      25.4 &
      20.8 &
      43.0 \\
     &
      Full (epochs=4) &
      \textbf{41.2} &
      49.3 &
      54.6 &
      59.4 &
      \textbf{48.1} &
      \textbf{30.1} &
      19.6 &
      43.2 \\ \bottomrule
    \end{tabular}%
  }
  \caption{
  Complete results for Table \ref{tab:main_results-mistral_select} in the main text, 
  with \textbf{Mistral-7B-v0.3} as the base model.
  Under all three budgets, \name still maintains the strongest macro-average performance, and ranks either first or second among the six candidate selection methods on 5/7 benchmarks on average. 
  The top 15\% BIDS-selected subset again outperforms full dataset training in macro average, by steadily improving on coding and math while maintaining its remarkable general instruction-following ability.}
  \label{tab:append_results-mistral_select-non_influence}
\end{table*}

\begin{table*}[!htb]
  \centering
  \resizebox{\linewidth}{!}{
  \begin{tabular}{@{}cccccccccc@{}}
    \toprule
    \multirow{2}{*}{\textbf{Budget}} &
      \multirow{2}{*}{\textbf{Method}} &
      \multicolumn{2}{c}{\textbf{Coding}} &
      \textbf{Logic} &
      \textbf{Knowledge} &
      \multicolumn{2}{c}{\textbf{Math}} &
      \textbf{Ins-Following} &
      \multirow{2}{*}{\textbf{Macro Avg}} \\ \cmidrule(lr){3-9}
     &
       &
      HumanEval &
      MBPP &
      BBH &
      MMLU &
      GSM-Plus &
      MATH &
      IFEval &
       \\ \midrule
    \multirow{6}{*}{5\%} &
      Task-max (LESS) &
      {\ul 36.8} &
      \textbf{45.6} &
      \textbf{60.1} &
      \textbf{62.3} &
      \textbf{38.6} &
      19.5 &
      23.8 &
      \textbf{41.0} \\
     &
      Sum &
      35.1 &
      43.3 &
      59.7 &
      {\ul 62.1} &
      {\ul 38.2} &
      17.6 &
      23.2 &
      39.9 \\
     &
      Temperature &
      34.2 &
      42.4 &
      {\ul 59.9} &
      61.8 &
      37.2 &
      {\ul 19.7} &
      21.8 &
      39.6 \\
     &
      \textbf{BIDS (ours)} &
      \textbf{37.7} &
      44.4 &
      59.5 &
      61.8 &
      38.0 &
      \textbf{19.8} &
      {\ul 26.1} &
      \textbf{41.0} \\
     &
      $- \texttt{Iter}$ &
      {\ul 36.8} &
      44.1 &
      59.1 &
      61.5 &
      {\ul 38.2} &
      19.6 &
      \textbf{27.5} &
      \textbf{41.0} \\
     &
      $- (\texttt{Norm} + \texttt{Iter})$ &
      33.3 &
      {\ul 45.0} &
      59.3 &
      61.6 &
      38.0 &
      18.7 &
      22.0 &
      39.7 \\ \midrule
    \multirow{6}{*}{10\%} &
      Task-max (LESS) &
      {\ul 39.5} &
      43.8 &
      {\ul 60.3} &
      \textbf{62.0} &
      39.7 &
      20.0 &
      24.6 &
      41.4 \\
     &
      Sum &
      36.8 &
      43.0 &
      59.7 &
      61.6 &
      40.1 &
      18.0 &
      22.0 &
      40.2 \\
     &
      Temperature &
      35.1 &
      {\ul 45.6} &
      60.0 &
      \textbf{62.0} &
      39.1 &
      18.4 &
      22.2 &
      40.3 \\
     &
      \textbf{BIDS (ours)} &
      \textbf{40.4} &
      \textbf{46.1} &
      \textbf{60.5} &
      61.7 &
      {\ul 40.5} &
      \textbf{21.0} &
      \textbf{27.1} &
      \textbf{42.5} \\
     &
      $- \texttt{Iter}$ &
      37.7 &
      45.0 &
      59.7 &
      61.6 &
      40.2 &
      20.2 &
      {\ul 26.7} &
      {\ul 41.6} \\
     &
      $- (\texttt{Norm} + \texttt{Iter})$ &
      36.0 &
      43.8 &
      59.7 &
      61.5 &
      \textbf{41.6} &
      {\ul 20.8} &
      24.6 &
      41.1 \\ \midrule
    \multirow{6}{*}{15\%} &
      Task-max (LESS) &
      \textbf{40.4} &
      44.1 &
      59.6 &
      61.3 &
      {\ul 41.9} &
      {\ul 21.5} &
      24.0 &
      41.8 \\
     &
      Sum &
      37.7 &
      {\ul 47.0} &
      59.4 &
      61.3 &
      41.2 &
      18.1 &
      23.8 &
      41.2 \\
     &
      Temperature &
      37.7 &
      \textbf{47.9} &
      \textbf{60.5} &
      \textbf{61.7} &
      39.7 &
      18.4 &
      24.2 &
      41.4 \\
     &
      \textbf{BIDS (ours)} &
      {\ul 39.5} &
      {\ul 47.0} &
      59.5 &
      \textbf{61.7} &
      41.8 &
      \textbf{21.7} &
      \textbf{27.1} &
      \textbf{42.6} \\
     &
      $- \texttt{Iter}$ &
      38.6 &
      46.4 &
      59.8 &
      61.6 &
      41.8 &
      20.9 &
      {\ul 26.1} &
      {\ul 42.2} \\
     &
      $- (\texttt{Norm} + \texttt{Iter})$ &
      {\ul 39.5} &
      45.0 &
      {\ul 60.1} &
      61.6 &
      \textbf{42.8} &
      {\ul 21.5} &
      23.2 &
      42.0 \\ \bottomrule
    \end{tabular}%
  }
  \caption{
  Complete results for Table \ref{tab:main_results-influence} and \ref{tab:sep_contrib} in the main text, 
  with \textbf{Mistral-7B-v0.3} as the base model.
  Performance improvements from \textbf{$- (\texttt{Norm} + \texttt{Iter})$} to \textbf{$- \texttt{Iter}$} to \name are consistent with Llama-3-8B in 
  \protect\S\ref{sec:ablation}. 
  }
  \label{tab:append_results-mistral_select-influence}
\end{table*}

\begin{figure*}[!htb]
  \centering
  \begin{minipage}{.48\textwidth}
    \input{figures/mistral_select/AID_all-mistral_select}
  \end{minipage}
  \hspace{0.02\textwidth} %
  \begin{minipage}{.48\textwidth}
    \input{figures/mistral_select/THI_unnorm_task_p0.1-mistral_select}
  \end{minipage}
\end{figure*}

\begin{figure*}[!htb]
    \centering
    \begin{subfigure}{0.32\textwidth}        
        \input{figures/mistral_select/THI_unnorm_col_p0.1-mistral_select}
    \end{subfigure}
    \hfill %
    \begin{subfigure}{0.32\textwidth}
        \input{figures/mistral_select/THI_norm_col_p0.1-mistral_select}
    \end{subfigure}
    \hfill %
    \begin{subfigure}{0.32\textwidth}
        \input{figures/mistral_select/THI_BIDS_p0.1-mistral_select}
    \end{subfigure}
    \caption{THI under the 10\% budget, with the base model being Mistral-7B-v0.3. Similar to prior observations with Llama-3-8B, both \textbf{$- \texttt{Iter}$} and \name have more balanced task frequencies than \textbf{$- (\texttt{Norm} + \texttt{Iter})$}.}
    \label{fig:compare_THI-mistral_select}
\end{figure*}

\begin{figure*}[!htb]
    \centering
    \begin{subfigure}{0.32\textwidth}
        
        \input{figures/mistral_select/AID_unnorm_col_p0.1-mistral_select}
    \end{subfigure}
    \hfill %
    \begin{subfigure}{0.32\textwidth}

        \input{figures/mistral_select/AID_norm_col_p0.1-mistral_select}
    \end{subfigure}
    \hfill %
    \begin{subfigure}{0.32\textwidth}

        \input{figures/mistral_select/AID_BIDS_p0.1-mistral_select}
    \end{subfigure}
    \caption{Comparative analysis of normalized AID under the 10\% budget, with the base model being Mistral-7B-v0.3. Similar to prior observations with Llama-3-8B, from \textbf{$- (\texttt{Norm} + \texttt{Iter})$} to \textbf{$- \texttt{Iter}$} to \name, the disparity among different tasks and instances in AID gradually diminishes, with both decreasing maximums and increasing minimums, although the degree of the original imbalance for Mistral-v0.3 is not as high as Llama-3.}
    \label{fig:compare_AID-mistral_select}
\end{figure*}

Similar to the analysis framework in 
\S\ref{sec:analysis}, 
we also present the AID analysis of the whole UltraInteract dataset (Figure \ref{fig:AID_all-mistral_select}) and the THI analysis of LESS-selected data (Figure \ref{fig:THI_unnorm_task_p0.1-mistral_select}). Then we follow the workflow in  \S\ref{sec:ablation}
to present both THI and AID anlyses for the three progressive algorithms: \textbf{$- (\texttt{Norm} + \texttt{Iter})$}, \textbf{$- \texttt{Iter}$} and \name (Figure \ref{fig:compare_THI-mistral_select}, \ref{fig:compare_AID-mistral_select}). The only difference here is that the selection model is Mistral-7B-v0.3 instead of Llama-3-8B.

\subsection{Applying \name to Other Data Selection Methods than LESS}\label{append:7_RDS_BIDS}

\begin{table*}[!ht]
  \centering
  \resizebox{\linewidth}{!}{
  \begin{tabular}{@{}cccccccccc@{}}
    \toprule
    \multirow{2}{*}{\textbf{Budget}} &
      \multirow{2}{*}{\textbf{Method}} &
      \multicolumn{2}{c}{\textbf{Coding}} &
      \textbf{Logic} &
      \textbf{Knowledge} &
      \multicolumn{2}{c}{\textbf{Math}} &
      \textbf{Ins-Following} &
      \multirow{2}{*}{\textbf{Macro Avg}} \\ \cmidrule(lr){3-9}
                          &                   & HumanEval & MBPP & BBH  & MMLU & GSM-Plus & MATH & IFEval &               \\ \midrule
    \multirow{3}{*}{5\%}  & Random            & 43.5      & 48.9 & 64.8 & 64.9 & 41.5     & 22.5 & 18.1   & 43.4          \\
                          & RDS               & 45.6      & 52.7 & 62.2 & 65.0 & 34.5     & 17.2 & 15.5   & 41.8          \\
                          & \textbf{RDS-BIDS} & 50.9      & 53.0 & 64.1 & 65.0 & 41.0     & 19.7 & 20.0   & \textbf{44.8} \\ \midrule
    \multirow{3}{*}{10\%} & Random            & 47.8      & 50.6 & 65.0 & 64.9 & 43.9     & 24.0 & 17.8   & 44.9          \\
                          & RDS               & 50.0      & 54.7 & 63.2 & 64.6 & 39.3     & 22.4 & 18.3   & 44.6          \\
                          & \textbf{RDS-BIDS} & 47.4      & 53.6 & 63.2 & 65.2 & 43.6     & 24.1 & 21.6   & \textbf{45.5} \\ \midrule
    \multirow{3}{*}{15\%} & Random            & 48.7      & 51.9 & 65.2 & 65.1 & 45.6     & 25.0 & 18.8   & 45.7          \\
                          & RDS               & 50.0      & 53.9 & 63.7 & 64.5 & 41.1     & 23.5 & 18.1   & 45.0          \\
                          & \textbf{RDS-BIDS} & 48.2      & 53.0 & 63.8 & 64.8 & 45.1     & 24.2 & 21.4   & \textbf{45.8} \\ \bottomrule
    \end{tabular}%
  }
  \caption{
  When the key technical components of \name (i.e., (1) instance-level normalization, and (2) iterative selection favoring underrepresented tasks) are applied to the utility matrix of RDS, the resulting method, RDS-\name, significantly improves over the original RDS and consistently outperforms the random baseline. These results are obtained with Llama-3-8B as the base model.
  }
  \label{tab:append_results-llama_select-RDS_BIDS}
\end{table*}

Theoretically, \name is applicable to any data selection method that requires a validation set and computes a score between each pair of training and validation examples (i.e., produces a matrix $A$ where element $A_{ij}$ indicates the utility of $i$-th training example on $j$-th validation instance). We empirically apply the two techniques of \name, instance-level normalization and iterative selection favoring underrepresented tasks, to RDS, which computes pairwise scores using the cosine similarity between hidden-layer representations of training and validation examples. The resulting method is dubbed ``RDS-\name''. This attempt is motivated by the observations in Table \ref{tab:main_results-llama_select} that RDS-selected data also demonstrate significant bias towards coding-related tasks and lead to overall poor performance. We maintain the same training and evaluation setup, and show the comparison results across three budgets in Table \ref{tab:append_results-llama_select-RDS_BIDS}. 

We can see that the balancing techniques of \name work well for RDS-computed scores. Across all three budgets, RDS-\name significantly improves over the original RDS and consistently outperforms the random baseline. In terms of task-specific performance, RDS-\name also demonstrates extensive and balanced improvements in tasks other than coding, especially math and instruction-following. This shows the necessity of balancing various types of utility scores in general-purpose instruction tuning, in response to the Limitations (\S\ref{limitations}) section in the main text, and validates the broad applicability of the underlying principles of \name. 
However, as also stated in \S\ref{limitations}, the focus of our work on influence-based approaches limits detailed investigation into the more general imbalance of data selection scores under a multi-capability learning setup. We thus leave further exploration in this direction to future work.

\subsection{Generalizability of \name to Different Evaluation Suites, Training Datasets, and Model Sizes}\label{append:8_generalize}

\begin{table*}[!ht]
  \centering
  \resizebox{\linewidth}{!}{
  \begin{tabular}{@{}ccccccc@{}}
    \toprule
    \multirow{2}{*}{\textbf{Method}} & \multicolumn{2}{c}{\textbf{Coding}} & \multicolumn{2}{c}{\textbf{Math}} & \textbf{Ins-Following} & \multirow{2}{*}{\textbf{Macro Avg}} \\ \cmidrule(lr){2-6}
                    & HumanEval     & MBPP          & GSM-Plus      & MATH          & IFEval        &               \\ \midrule
    Random          & \textbf{47.8} & \textbf{50.6} & 43.9          & 24.0          & 17.8          & 36.8          \\
    Task-max (LESS) & 46.5          & 49.0          & 43.0          & 23.9          & 19.3          & 36.3          \\
    \textbf{BIDS}   & 46.5          & 49.9          & \textbf{44.5} & \textbf{25.3} & \textbf{24.6} & \textbf{38.2} \\ \bottomrule
    \end{tabular}%
  }
  \caption{
  Results of \textbf{Llama-3-8B}, with the training set being UltraInteract and the selection budget being 10\%. \textbf{MMLU and BBH, the two tasks with highest and lowest average influence respectively, are dropped from the original evaluation suite.} 
  This validates that the effectiveness of \name is not a special case under certain task setup, and can generalize to different evaluation suites.
  }
  \label{tab:append_results-llama_select-diff_eval}
\end{table*}

\begin{table*}[!ht]
  \centering
  \resizebox{\linewidth}{!}{
  \begin{tabular}{@{}ccccccc@{}}
    \toprule
    \multirow{2}{*}{\textbf{Method}} & \textbf{Coding} & \textbf{Logic} & \textbf{Knowledge} & \textbf{Math} & \textbf{Ins-Following} & \multirow{2}{*}{\textbf{Macro Avg}} \\ \cmidrule(lr){2-6}
                    & HumanEval     & BBH           & MMLU          & GSM-Plus      & IFEval        &               \\ \midrule
    Random          & 40.4          & 40.0          & \textbf{65.5} & 12.2          & 33.8          & 38.4          \\
    Task-max (LESS) & 44.7          & 24.3          & 65.0          & \textbf{16.9} & 38.1          & 37.8          \\
    \textbf{BIDS}   & \textbf{45.6} & \textbf{53.5} & 64.7          & \textbf{16.9} & \textbf{41.3} & \textbf{44.4} \\ \bottomrule
    \end{tabular}%
  }
  \caption{
  Results of \textbf{Llama-3-8B}, with the training set being the original mixture used in \citet{xia2024less} (\cott ~\citep{wei2022chain}, \dolly ~\citep{DatabricksBlog2023DollyV2}, \flan ~\citep{longpre2023flan}, and \oasst ~\citep{kopf2023openassistant}), and the selection budget being 10\%. 
  This validates that the effectiveness of \name can generalize to different training datasets. 
  }
  \label{tab:append_results-llama_select-diff_training}
\end{table*}

\begin{table*}[!ht]
  \centering
  \resizebox{\linewidth}{!}{
  \begin{tabular}{@{}ccccccccc@{}}
    \toprule
    \multirow{2}{*}{\textbf{Method}} &
      \multicolumn{2}{c}{\textbf{Coding}} &
      \textbf{Logic} &
      \textbf{Knowledge} &
      \multicolumn{2}{c}{\textbf{Math}} &
      \textbf{Ins-Following} &
      \multirow{2}{*}{\textbf{Macro Avg}} \\ \cmidrule(lr){2-8}
                    & HumanEval & MBPP & BBH           & MMLU & GSM-Plus & MATH          & IFEval &      \\ \midrule
    Random          & 30.7      & 39.5 & \textbf{48.6} & 56.6 & 25.2     & \textbf{14.1} & 12.0   & 32.4 \\
    Task-max (LESS) & 28.9      & 40.7 & 46.9          & 56.1 & 23.6     & 13.5          & 14.7   & 32.1 \\
    \textbf{BIDS} &
      \textbf{33.3} &
      \textbf{42.1} &
      48.2 &
      \textbf{56.8} &
      \textbf{25.7} &
      \textbf{14.1} &
      \textbf{17.1} &
      \textbf{33.9} \\ \bottomrule
    \end{tabular}%
  }
  \caption{
  Results of \textbf{Llama-3.2-3B}, with the training set being UltraInteract and the selection budget being 10\%. This validates that the effectiveness of \name can generalize to models of different sizes. 
  }
  \label{tab:append_results-llama_select-diff_model_size}
\end{table*}

In this section, we additionally validate the effectiveness of \name under three different experimental setups that vary in evaluation suites, training datasets, and size of the base model. Results in this section show that the effectiveness of \name is robust and generalizable across different experimental configurations, thus further solidifying the claims in the main text.

\paragraph{Different evaluation suites.} 
First, we keep the same training dataset (i.e., UltraInteract), but drop MMLU and BBH, the two tasks with highest and lowest average influence respectively, from the evaluation suite. In this way we investigate how \name performs when the disparity of cross-task influence decreases. This evaluation setup also aligns more with the practical use case where an AI agent should not only have strong math and coding capabilities but also faithfully follow user instructions.
From the results in Table~\ref{tab:append_results-llama_select-diff_eval}, we find that despite decreased cross-task influence discrepancies, \name is still effective in balancing and improving performance across tasks. Notably, it advances the performance of the original LESS-selected data in all 5 evaluation tasks.

\paragraph{Different training datasets.}
Next, we investigate how \name performs with a different training dataset. For better comparison with prior work, we adopt the original training mixture used in LESS, which is composed of \cott ~\citep{wei2022chain}, \dolly ~\citep{DatabricksBlog2023DollyV2}, \flan ~\citep{longpre2023flan}, and \oasst ~\citep{kopf2023openassistant}, and has 270K examples altogether. 
This training dataset is more general-purposed compared with the reasoning focus of UltraInteract, thus suitable for testing the universal applicability of \name. For the evaluation suite, we drop MBPP and MATH, and evaluate on the remaining five tasks to align with the general property of the new training dataset.
Table~\ref{tab:append_results-llama_select-diff_training} still shows significant improvements of \name over other baselines. Notably, we observe the similar counterintuitive phenomenon for LESS to \S\ref{sec:analysis}, where MMLU has the highest average influence but does not show greater room for performance improvements. We also investigate why LESS shows extremely poor performance in BBH. It turns out that models trained on LESS-selected data tend to stop generation right after generating ``Let’s think step by step'', which is likely a consequence of training on imbalanced data mixtures.

\paragraph{Different size of the base model.}
Moreover, to verify that the effectiveness of \name is robust to the size of the base model, we experiment with Llama-3.2-3B using the same training and evaluation setup as the Llama-3-8B and Mistral-7B-v0.3 experiments in the main text.
Table~\ref{tab:append_results-llama_select-diff_model_size} again validates that the effectiveness of \name can generalize to modern smaller models. Notably, apart from improved average performance, \name also outperforms the random baseline and the original LESS in 6/7 and 7/7 tasks respectively, demonstrating the generalizability of its balancing effect.

\subsection{Discussion on the Computational Cost}\label{append:8_cost}

In this section, we aim to discuss and show that \name does not incur much memory or latency overhead, and can thus serve as an efficient plug-and-play module. In our training and evaluation setup, the $|D|$ dimension for the Attribution Matrix is about 288K~\citep{yuan2024advancing}, and the $|V|$ dimension is 350. Therefore, the memory cost for storing the AM using FP64 precision is less than 800 MB. The latency cost for running the whole \name algorithm is less than 1 minute with CUDA acceleration of a single H100 GPU. More generally, since many popular mixtures of instruction finetuning data are maintained on the scale of hundreds of thousands~\citep{wang2023far, ivison2023camels, yuan2024advancing, yue2023mammoth}, the memory and latency cost of BIDS should be light for most practical training setups.

\subsection{Qualitative Analysis}\label{append:9_examples}

In this section, we aim to demonstrate the following two properties of \name with some qualitative examples, and better illustrate its effectiveness:
\begin{enumerate}
    \item Models trained on \name-selected data can indeed achieve a stronger balance between mastering task-specific skills (e.g., math reasoning, coding knowledge, etc.) and fully understanding various types of instructions given by the user (e.g., format-following, response style, etc.).
    \item Such a stronger balance is indeed helpful to improving the accuracy or human-perceived quality of model response.
\end{enumerate}

Concretely, we present three sets of model responses in the task of coding (Table~\ref{tab:qual-coding}), math (Table~\ref{tab:qual-math}) and general instruction-following (Table~\ref{tab:qual-if}) respectively. Each set contains a correct response by a Mistral-7B-v0.3 model trained on top-15\% \name-selected data, and a false response by the same base model trained on the full (i.e., 100\%) UltraInteract, both to exactly the same prompt. We analyze how the \name-trained model correctly answers all these prompts due to the greater balance of capabilities it achieves.

\begin{table*}[htbp]
  \centering
  \caption{For the example 1, the model trained on the full dataset fails to handle the corner case of \texttt{numbers = []}. For the example 2, the full-trained model also fails at not adding the constraint of \texttt{y != x} in its sorting rule. In both cases, \name-trained model returns the correct code completion because it better considers and handles corner cases. It reflects that \name-trained model balances its capability in correct coding knowledge and comprehensive thinking behavior.}
  \renewcommand{\arraystretch}{1.5} %
  \adjustbox{max width=\textwidth}{
    \begin{tabular}{p{26em}p{26em}}
    \toprule
    \textbf{HumanEval: Example 1} & \textbf{HumanEval: Example 2} \\
    \cmidrule(lr){1-1} \cmidrule(lr){2-2}
    \multicolumn{1}{l}{\textbf{Prompt}} & \multicolumn{1}{l}{\textbf{Prompt}} \\
    \cmidrule(lr){1-1} \cmidrule(lr){2-2}
\begin{lstlisting}[numbers=none]
def rolling_max(numbers):
    # Write a Python function `rolling_max(numbers: List[int]) -> List[int]` to solve the following problem:
    # From a given list of integers, generate a list of rolling maximum element found until given moment
    # in the sequence.
    # >>> rolling_max([1, 2, 3, 2, 3, 4, 2])
    # [1, 2, 3, 3, 3, 4, 4]
    
\end{lstlisting} & 
\begin{lstlisting}[numbers=none]
def find_closest_elements(numbers: List[float]) -> Tuple[float, float]
    # From a supplied list of numbers (of length at least two) select and return two that are the closest to each other and return them in order (smaller number, larger number).
    # >>> find_closest_elements([1.0, 2.0, 3.0, 4.0, 5.0, 2.2])
    # (2.0, 2.2)
    # >>> find_closest_elements([1.0, 2.0, 3.0, 4.0, 5.0, 2.0])
    # (2.0, 2.0)
    
\end{lstlisting} \\
    \cmidrule(lr){1-1} \cmidrule(lr){2-2}
    \multicolumn{1}{l}{\textbf{Wrong Completion: Full-trained model}} & \multicolumn{1}{l}{\textbf{Wrong Completion: Full-trained model}} \\
    \cmidrule(lr){1-1} \cmidrule(lr){2-2}
\begin{lstlisting}[numbers=none]
# Initialize the result list with the first element of the input list
@result = [numbers[0]]@
# Initialize the maximum value with the first element of the input list
max_value = numbers[0]
# Iterate through the input list starting from the second element
for num in numbers[1:]:
    # If the current number is greater than the maximum value, update the maximum value and append the current number to the result list
    if num > max_value:
        max_value = num
        result.append(num)
    # If the current number is not greater than the maximum value, append the maximum value to the result list
    else:
        result.append(max_value)
# Return the result list
return result
\end{lstlisting} & 
\begin{lstlisting}[numbers=none]
@closest_pair = min(numbers, key=lambda x: min(abs(x - y) for y in numbers))@
return closest_pair, min(numbers, key=lambda x: abs(x - closest_pair))
\end{lstlisting} \\
    \cmidrule(lr){1-1} \cmidrule(lr){2-2}
    \multicolumn{1}{l}{\textbf{Correct Completion: \name-trained model}} & \multicolumn{1}{l}{\textbf{Correct Completion: \name-trained model}} \\
    \cmidrule(lr){1-1} \cmidrule(lr){2-2}
\begin{lstlisting}[numbers=none]
`if not numbers:`
    `return []`
n = len(numbers)
result = [numbers[0]]
for i in range(1, n):
    result.append(max(numbers[i], result[-1]))
return result
\end{lstlisting} & 
\begin{lstlisting}[numbers=none]
# Step 1: Sort the list of numbers in ascending order
`numbers.sort()`
# Step 2: Initialize the closest_numbers variable with the first two elements of the sorted list
closest_numbers = (numbers[0], numbers[1])
# Step 3: Iterate through the list to find the two numbers that are the closest to each other
for i in range(1, len(numbers) - 1):
    # Step 4: Calculate the difference between the current number and the next number
    diff = abs(numbers[i] - numbers[i + 1])
    # Step 5: If the difference is less than the difference between the current closest numbers, update the closest numbers
    if diff < abs(closest_numbers[0] - closest_numbers[1]):
        closest_numbers = (numbers[i], numbers[i + 1])
# Step 6: Return the two closest numbers in order (smaller number, larger number)
return closest_numbers
\end{lstlisting} \\
    \bottomrule
    \end{tabular}%
  }
  \label{tab:qual-coding}%
\end{table*}%

\begin{table*}[htbp]
  \centering
  \caption{For the example 1, \name-trained model understands the implicit condition contained in the question that only after a whole number of years can a tree grow oranges for sale, and thus goes on to round the calculated the calculation to integer (Step 7). For the example 2, the model trained on the full dataset fails to understand the meaning of "a person was born 8 years before another" (Step 3) and thus performs the calculation in the wrong way. In both examples, \name-trained model correctly answers the question because it balances its capability in math reasoning and natural language understanding.}
  \renewcommand{\arraystretch}{1.5} %
  \adjustbox{max width=\linewidth}{
  \Large
    \begin{tabular}{p{26em}p{24em}}
    \toprule
    \multicolumn{1}{l}{\textbf{GSM-Plus: Example 1}} & \multicolumn{1}{l}{\textbf{GSM-Plus: Example 2}} \\
    \cmidrule(lr){1-1} \cmidrule(lr){2-2}
    \multicolumn{1}{l}{\textbf{Question}} & \multicolumn{1}{l}{\textbf{Question}} \\
    \cmidrule(lr){1-1} \cmidrule(lr){2-2}
    Solve the following math problem step-by-step.\newline{}Simplify your answer as much as possible. Present your final answer as \textbackslash{}boxed\{Your Answer\}.\newline{}Carlos is planting an orange tree. The tree will cost \$120 to plant. Each year it will grow 10 oranges, which he can sell for \$2 each. It costs \$4 a year to water and feed the tree. How many years will it take before he starts earning money on the orange tree? & Solve the following math problem step-by-step.\newline{}Simplify your answer as much as possible. Present your final answer as \textbackslash{}boxed\{Your Answer\}.\newline{} Raymond and Samantha are cousins. Raymond was born 8 years before Samantha. Raymond had a son at the age of 25. If Samantha is now 35, how many years ago was Raymond's son born? \\
    \cmidrule(lr){1-1} \cmidrule(lr){2-2}
    \multicolumn{1}{l}{\textbf{Wrong Response: Full-trained model}} & \multicolumn{1}{l}{\textbf{Wrong Response: Full-trained model}} \\
    \cmidrule(lr){1-1} \cmidrule(lr){2-2}
    Step 1:  To determine the number of years it will take for Carlos to start earning money from the orange tree, we need to consider the initial cost of planting the tree and the annual expenses associated with watering and feeding it.\newline{}Step 2:  The tree costs \$120 to plant, and it incurs an annual expense of \$4 for watering and feeding.\newline{}Step 3:  Each year, the tree produces 10 oranges, which he can sell for \$2 each.\newline{}Step 4:  So, the total income from selling the oranges each year is 10 * \$2 = \$20.\newline{}Step 5:  However, we need to subtract the annual expenses of \$4 from the income to find the net profit from the oranges.\newline{}Step 6:  Therefore, the net profit from the oranges each year is \$20 - \$4 = \$16.\newline{}Step 7:  To determine the number of years it will take for Carlos to start earning money, we can divide the initial cost of planting the tree by the net profit from the oranges each year.\newline{}Step 8:  This gives us the number of years it will take for the tree to start generating a positive return on investment.\newline{}Step 9:  So, the number of years it will take before Carlos starts earning money on the orange tree is \$120 / \$16 = 7.5\$ years.\newline{}\newline{}Answer:\newline{}\textbackslash{}boxed\{7.5\} & Step 1:  To solve this problem, we need to first understand the given information and then work through the steps to find the age of Raymond's son.\newline{}Step 2:  We're given that Raymond was born 8 years before Samantha, and Samantha is currently 35 years old.\newline{}\textcolor{red}{Step 3:  This means that Raymond was born 35 - 8 = 27 years ago.}\newline{}Step 4:  Now, we know that Raymond had a son at the age of 25. So, we need to find out how many years ago Raymond's son was born.\newline{}Step 5:  To do this, we can simply subtract Raymond's age when his son was born (27 years) from the current age of Samantha (35 years).\newline{}Step 6:  This gives us 35 - 27 = 8 years.\newline{}Step 7:  Therefore, Raymond's son was born 8 years ago.\newline{}\newline{}Answer:\newline{}\textbackslash{}boxed\{8\} \\
    \cmidrule(lr){1-1} \cmidrule(lr){2-2}
    \multicolumn{1}{l}{\textbf{Correct Response: \name-trained model}} & \multicolumn{1}{l}{\textbf{Correct Response: \name-trained model}} \\
    \cmidrule(lr){1-1} \cmidrule(lr){2-2}
    Step 1:  To determine how many years it will take for Carlos to start earning money from his orange tree, we need to consider the costs involved in planting and maintaining the tree, as well as the revenue generated from selling the oranges.\newline{}Step 2:  First, we need to calculate the total cost of planting the tree, which is \$120.\newline{}Step 3:  Each year, the tree will grow 10 oranges that can be sold for \$2 each, resulting in a total revenue of \$20 per year.\newline{}Step 4:  However, it also costs \$4 per year to water and feed the tree.\newline{}Step 5:  Therefore, the net revenue from selling the oranges each year is \$20 - \$4 = \$16.\newline{}Step 6:  To determine how many years it will take for Carlos to start earning money, we can divide the total cost of planting the tree by the net revenue each year. This gives us 120 / 16 = 7.5 years.\newline{}\textcolor{mdgreen}{Step 7:  Since we can't plant a tree for half a year, Carlos will need to wait 8 years before he starts earning money from the orange tree.}\newline{}\newline{}Answer:\newline{}\textbackslash{}boxed\{8\} & Step 1:  To solve this problem, we need to understand the relationship between Raymond and Samantha, and then calculate the age of Raymond's son based on the given information.\newline{}Step 2:  Raymond is Samantha's cousin, and he was born 8 years before her. This means that Raymond is 8 years older than Samantha.\newline{}Step 3:  Raymond had a son at the age of 25, so we need to calculate Raymond's age at that time. Since Raymond is 8 years older than Samantha, and Samantha is currently 35, we can calculate Raymond's age as 35 + 8 = 43.\newline{}Step 4:  Now, we know that Raymond had his son at the age of 25, so we can calculate the age of Raymond's son by subtracting 25 from Raymond's age at that time, which is 43 - 25 = 18.\newline{}Step 5:  Therefore, Raymond's son was born 18 years ago.\newline{}\newline{}Answer:\newline{}\textbackslash{}boxed\{18\} \\
    \bottomrule
    \end{tabular}%
    }
  \label{tab:qual-math}%
\end{table*}%

\begin{table*}[htbp]
  \centering
  \caption{For the example 1, the model trained on the full dataset ignores the explicit requirement of forming its output as a single JSON block. It is especially overfit to the response format of \texttt{Step X: ...} which is frequently seen in the training data. In contrast, \name-trained model handles its CoT in a nifty way by expressing its thinking process in the values of a JSON block. For the example 2, full-trained model not only fails to completely follow the explicitly required format, but also hardly writes a poem, compared with the high-quality one generated by \name-trained model. In both examples, \name-trained model generates content with significantly higher quality due to the balance between strict format-following and flexible use of parametric knowledge.}
  \renewcommand{\arraystretch}{1.5} %
  \adjustbox{max width=\linewidth}{
  \Large
    \begin{tabular}{p{24em}p{24em}}
    \toprule
    \multicolumn{1}{l}{\textbf{IFEval: Example 1}} & \multicolumn{1}{l}{\textbf{IFEval: Example 2}} \\
    \cmidrule(lr){1-1} \cmidrule(lr){2-2}
    \multicolumn{1}{l}{\textbf{Prompt}} & \multicolumn{1}{l}{\textbf{Prompt}} \\
    \cmidrule(lr){1-1} \cmidrule(lr){2-2}
    Please provide the names of 5 famous moms in JSON format. Please, use any interesting or weird tone. Your entire output should just contain a JSON block, nothing else. & Write a poem about how I am missing my classes. The poem must have 4 sections marked with SECTION X. Finish the poem with this exact phrase: "Can I get my money back for the classes I missed?" \\
    \cmidrule(lr){1-1} \cmidrule(lr){2-2}
    \multicolumn{1}{l}{\textbf{Wrong Response: Full-trained model}} & \multicolumn{1}{l}{\textbf{Wrong Response: Full-trained model}} \\
    \cmidrule(lr){1-1} \cmidrule(lr){2-2}
    Step 1: Let's think of 5 famous moms who have made a significant impact in their respective fields.\newline{}Step 2: Oprah Winfrey - Talk show host, media mogul, producer, actress, philanthropist\newline{}Step 3: Angelina Jolie - Actress, filmmaker, humanitarian\newline{}Step 4: Michelle Obama - Former First Lady of the United States\newline{}Step 5: Serena Williams - Professional tennis player\newline{}Step 6: Create a JSON block with the names of the 5 famous moms.\newline{}\newline{}\newline{}\verb|```|javascript\newline{}// Step 7: Create a JSON block with the names of the 5 famous moms.\newline{}[\newline{}  "Oprah Winfrey",\newline{}  "Angelina Jolie",\newline{}  "Michelle Obama",\newline{}  "Serena Williams"\newline{}]\newline{}\verb|```| & Section 1:\newline{}I missed my first class today and I'm feeling quite disappointed.\newline{}Section 2:\newline{}I really wanted to learn all the new material, but I had to stay home due to illness.\newline{}Section 3:\newline{}I hope I can make up for the missed class by studying extra hard for the next exam.\newline{}Section 4:\newline{}Can I get my money back for the classes I missed?\newline{}\newline{}Answer:\newline{}"I missed my first class today and I'm feeling quite disappointed.\newline{}I really wanted to learn all the new material, but I had to stay home due to illness.\newline{}I hope I can make up for the missed class by studying extra hard for the next exam.\newline{}Can I get my money back for the classes I missed?" \\
    \cmidrule(lr){1-1} \cmidrule(lr){2-2}
    \multicolumn{1}{l}{\textbf{Correct Response: \name-trained model}} & \multicolumn{1}{l}{\textbf{Correct Response: \name-trained model}} \\
    \cmidrule(lr){1-1} \cmidrule(lr){2-2}
    \{ \newline{}  "Meryl Streep": "An American actress and filmmaker, known for her versatile and powerful performances in films like Kramer vs. Kramer, Sophie's Choice, The Devil Wears Prada, and many more.",\newline{}  "Oprah Winfrey": "An American media mogul, talk show host, actress, producer, and philanthropist. She is best known for her talk show The Oprah Winfrey Show, which ran for 25 years.",\newline{}  "Sarah Jessica Parker": "An American actress, producer, and designer. She is best known for her role as Carrie Bradshaw in the HBO series Sex and the City.",\newline{}  "Jennifer Garner": "An American actress, producer, and entrepreneur. She is best known for her roles in films like 13 Going on 30, Juno, and Alias.",\newline{}  "Angelina Jolie": "An American actress, filmmaker, and humanitarian. She is best known for her work as a United Nations Special Envoy and as a Goodwill Ambassador."\newline{} \} & SECTION 1:\newline{}I'm sitting here feeling so lonely,\newline{}Missing my classes with all my friends,\newline{}Wishing I could be there with them,\newline{}Instead of sitting at home and mend.\newline{}SECTION 2:\newline{}I'm looking at my empty calendar,\newline{}Wondering what I'm missing out on,\newline{}Wishing I could join in on the fun,\newline{}Instead of staring at the screen and frown.\newline{}SECTION 3:\newline{}I'm listening to the sound of silence,\newline{}While my friends are making memories,\newline{}Wishing I could join in on the laughter,\newline{}Instead of feeling left out and sadder.\newline{}SECTION 4:\newline{}I'm staring at the clock, counting the minutes,\newline{}Until I can see my teachers and classmates,\newline{}Wishing I could just teleport there,\newline{}Instead of waiting for the next class to start.\newline{}\newline{}Finish:\newline{}Can I get my money back for the classes I missed? \\
    \bottomrule
    \end{tabular}%
    }
  \label{tab:qual-if}%
\end{table*}%

\end{document}